# Image Registration for the Alignment of Digitized Historical Documents

**AmirAbbas Davari**[1*], **Tobias Lindenberger**[1*], **Armin Häberle**[2], **Vincent Christlein**[1], **Andreas Maier**[1], **Christian Riess**[1]

1. Pattern Recognition Lab, Computer Science Department,
Friedrich-Alexander Universität Erlangen-Nürnberg, Erlangen, Germany
[amir.davari, tobias.lintob.lindenberger, vincent.christlein, andreas.maier, christian.riess] @fau.de

2. Bibliotheca Hertziana - Max-Planck-Institut für Kunstgeschichte, Rome, Italy
Haeberle@biblhertz.it

* Both authors contributed equally

## 1  Introduction

Novel imaging and image processing techniques provide technical tools to art historians for a better understanding of the creation of an artwork. The approach of a "work process analysis" of an artwork (e.g. a drawing or painting) by art historians aims at segmenting and differentiating the unique steps of production and thus, following the artist's path from the starting point to his final image. About 70-75% of all old master drawings consist of multiple materials, such as chalks of distinct colors, graphite and/or ink, which mostly have been applied in a step by step manner. This fact opens the opportunity for a chronological reconstruction of the genesis of the work. To this end, material decomposition of drawn layers is oftentimes the most accurate way to follow the artistic workflow.

One way to perform layer separation is by spectroscopy. However, this approach is oftentimes destructive to the examined material. To allow for an examination of old master drawings while preserving the drawing to the best extend possible, it is also possible to acquire a multi- or hyperspectral image of the drawing, and separate the layers within the range of visible wavelengths.

To obtain a reliable and consistent separation of artwork layers as a basis for art historical interpretation, a number of technical challenges have yet to be solved [Dav17]. First of all, there is the need for a pixel-wise "ground truth" map to objectively compare competing approaches. One possibility to get such a ground truth is to mimic the creation of a step-by-step layered artwork, and to image it after completing each work step. A map for a layer can then be obtained by subtracting two subsequent layers. However, one substantial issue lies in the fact that the acquisitions of two subsequent layers do not exactly match onto each other, and therefore have to be aligned.

There are many possible reasons for the mismatch, among which are distortions, mechanical motion, and spherical and chromatic aberration of the optical devices. An example mismatch is illustrated in Fig. 1.1. Similarly, a mismatch must be compensated when the output of an algorithm for layer separation should be mapped to the computed ground truth. This compensation must be a pixel-wise alignment. This is done by a process that is, in the field of image processing, referred to as "image registration".

In this work, we investigate different classical image registration methods for the purpose of creating an accurate ground truth map for hyperspectral historical document processing. We first narrow down the number of possibilities for solving this task by considering problem-specific constraints. Then, we quantitatively and qualitatively compare the two most promising approaches on a phantom document. This paper is organized as follows: in the main document, we state the key findings of this study. A more technical presentation and justification of the intermediate choices is provided in the appendix.

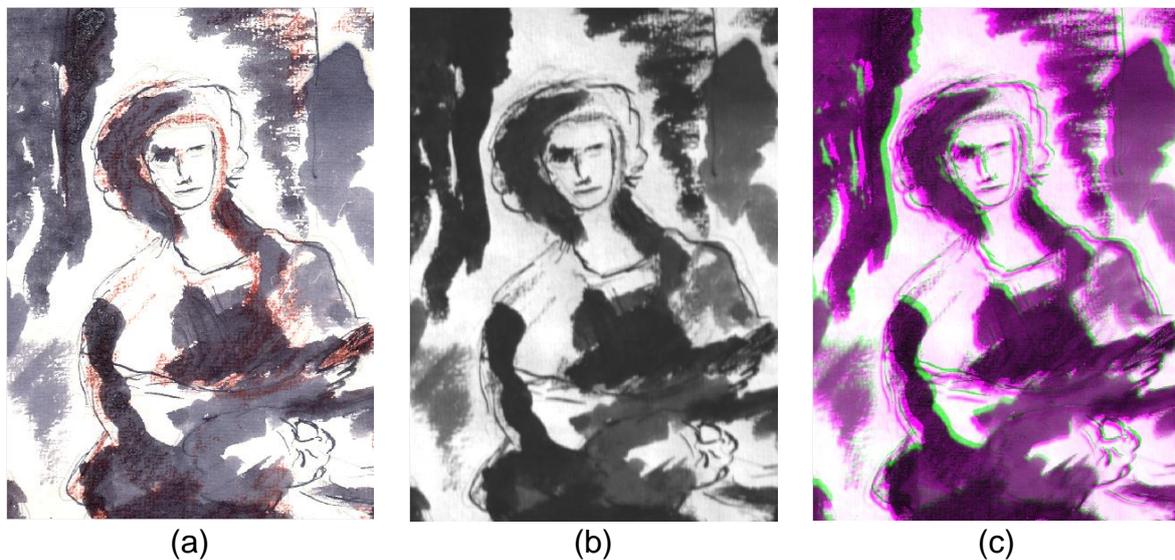

(a) (b) (c)

Figure 1.1: Importance of image registration for layer separation in old Master Drawings using image processing is depicted here. (a) image of a phantom data that is acquired by a board scanner, (b) sample channel of hyperspectral image from the same phantom data, (c) false color overlapping image of (a) and (b). As it can be observed, the two images that are acquired by the board scanner and the hyperspectral camera are not pixel-wise aligned. Therefore, the output of layer separation algorithm on the hyperspectral image cannot be numerically evaluated. Image registration would solve this problem.

## 2  Methods

### 2.1  Hyperspectral Image Acquisition

Hyperspectral imaging combines normal spatial imaging with spectroscopy. The spatial and spectral information of a target is stored in a stack of grayscale images. Each individual image in the stack represents the target recorded at a different wavelength. The stack can be used for further image analysis in the spatial and spectral domain [Mid16c].

For this project, we used a hyperspectral push-broom camera. This camera type features a two dimensional detector array that is combined with a spectrograph. The target is illuminated along one of the spatial axes of the detector. This line contains the full spectrum of the camera's spectral axis. In push-broom scanning, all image lines are imaged in order to acquire the full hyperspectral stack.

Further key components of the hyperspectral camera are the lens, the mirror scanner and the light source [Mid16b]. The data for this project was acquired with a Specim PFD-CL-65-V10E hyperspectral camera as it can be seen in Fig. 2.1. The camera is capable of recording 1040 spectral channels. Its sensor's spectral working range includes wavelengths between 400-1000 nanometers, i.e., ultraviolet (UV) to infrared (IR). Its spatial resolution is 1312 pixels. The camera sensor is manufactured in CCD technology and the setup works with a frame-rate up to 65 Hz [Mid16a].

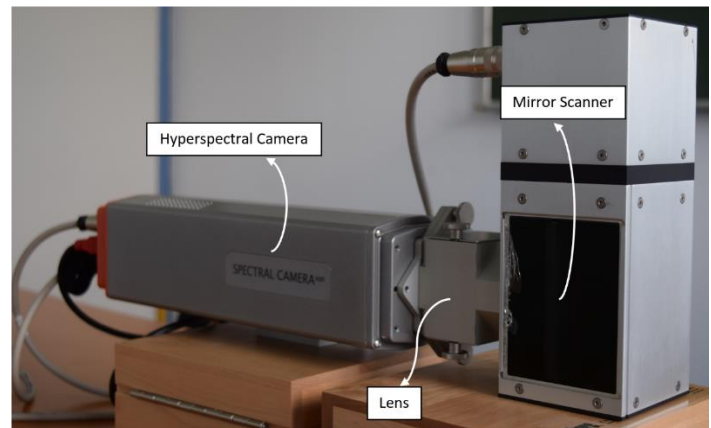

Figure 2.1: The hyperspectral camera that was used for data acquisition; a Specim PFD-CL-65-V10E hyperspectral camera.

## 2.2 Generation of Phantom Data

We created a phantom image with the goal of quantitatively evaluating layer separation algorithms. To design the phantom as close as possible to its historical counterpart, we used inks, chalks and papers that matched historical blueprints. To obtain maps of isolated layers, the image is drawn layer by layer, and scanned every time that a layer has been finalized. Three such example scans are shown in Fig. 2.2.

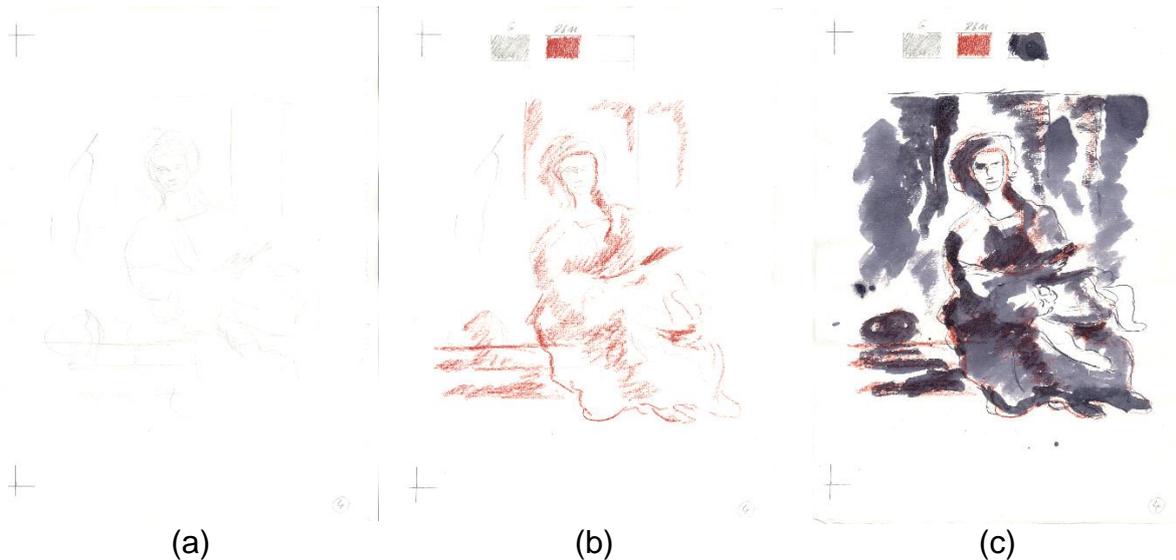

           (a)                                     (b)                                     (c)

Figure 2.2: The ground truth data. Layers of paint are consecutively added to the image. (a) first layer, (b) second layer, (c) third layer.

In this image series, a layer can be isolated by subtracting two subsequent images, but only if these two subsequent images are pixel-by-pixel aligned to each other. This alignment can be done in an automated way via image registration. Another registration step is required to evaluate the accuracy of a layer-separation algorithm on the hyperspectral image. In this case, the acquired hyperspectral image must be aligned pixel-by-pixel to the scanned ground truth phantom. In this work, we compare registration methods to solve the second registration task, i.e., registration between a hyperspectral acquisition of a drawing and a scanned ground truth image of a drawing.

## 2.3 Investigated Registration Approaches

There are various potential sources of distortions between the hyperspectral image and the scanned image, ranging from slight variations in document position, mechanical motions, to different lens effects. In many cases, the mismatch between both documents varies across the document area, and therefore requires a non-rigid registration.

A high-level picture of our workflow is depicted in Fig. 2.3. We first apply a coarse linear rescaling and alignment (pre-registration), and then apply Demon's registration algorithm [Thi98]. This method estimates a non-rigid displacement of each pixel based with the goal of maximizing the similarity of both documents. The definition of similarity is a key factor in the success of this algorithm. Thus, we considered several similarity measures, and out of these we selected specifically the localized mutual information (LMI) and the residual complexity (RC), as their mathematical properties are most promising for the task at hand. For optimization of the similarity function, we choose the steepest descent [Pre07], as it was also used in [Myr10]. For further details, please refer to the appendix, where we present the detailed results of our study.

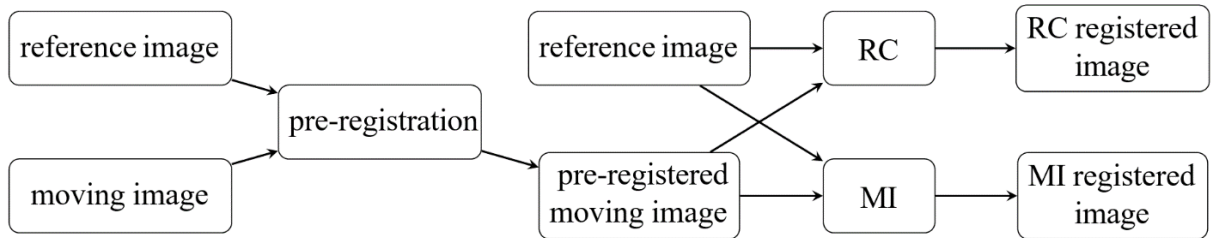

Figure 2.3: Flowchart for the complete registration procedure; first, a pre-registration is applied to the moving images; second, the pre-registered images are registered with the RC and MI algorithm respectively.

# 3 Evaluation

A common quantitative measure for registration performance is the Dice similarity coefficient (DSC) [Dic45]. It measures the regional overlap of segmented areas of both images. The value of DSC ranges between 0 and 1, with 0 indicating no overlap and 1 indicating a complete overlap of the segmented areas.

For the evaluation, we defined three regions of interest which are depicted in Fig. 3.1. First, we considered the full image area, second, we considered the face of the person as an example area that shows a lot of details, and third we considered the bottom part of the image for an example area showing only few details.

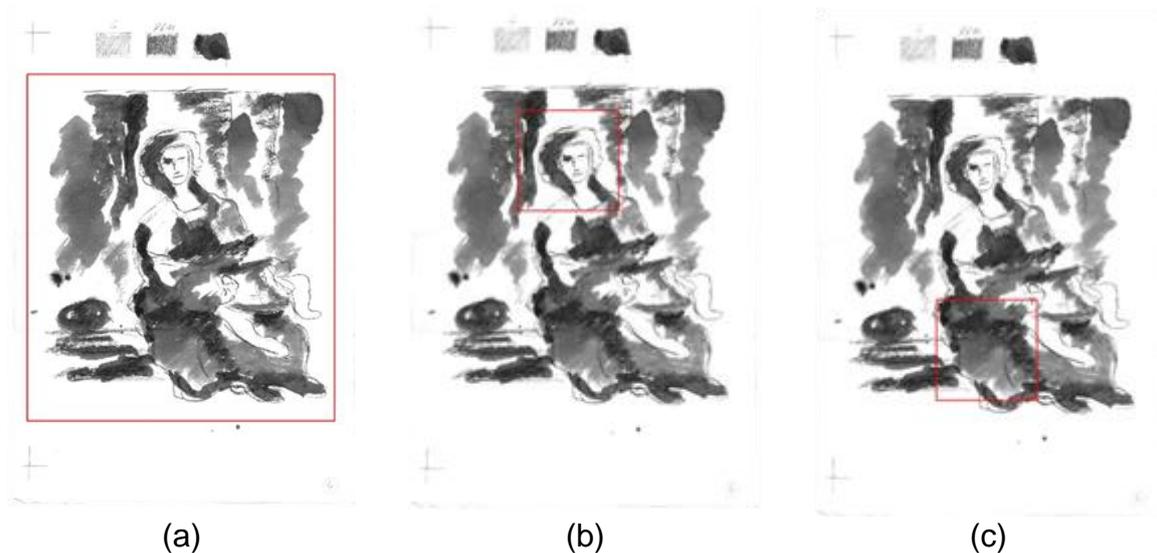

(a) (b) (c)

Figure 3.1: We defined three regions of interest for the evaluation purpose which are shown by a red rectangle. (a) Full image area, (b) high detail region, (c) low detail region.

Results for both the residual complexity (RC) and the localized mutual information (LMI) metric for these three regions are shown in Table 3.1. For the full area, RC and MI reach an equally high DSC score of 0.97. Both methods also provide good results in the high detail image region and the low detail image region, with no clear advantage for one method or another.

| Image | no registration | RC registered | MI registered |
|---|---|---|---|
| Full area | 0.75 | 0.97 | 0.97 |
| High details | 0.64 | 0.95 | 0.94 |
| Low details | 0.41 | 0.91 | 0.92 |

Table 3.1: DICE similarity coefficient (DSC) for the different regions of interest registration.

The results of high and low detail regions of interest are qualitatively depicted in Fig. 3.2 and Fig. 3.3, respectively. To make the visual inspection clearer, we computed the edge images of the two selected regions. The edges were extracted using the Sobel edge detector [Sob68].

As it is shown in Fig. 3.2 and 3.3, the edges that could be extracted from the images are not identical due to higher detail in the scanned images, noise in the registered hyperspectral images, and changes introduced to the images during registration. However, the salient structures that are present in the images could all be extracted. In Fig. 3.2, the structures of the face are clearly visible and in Fig. 3.3, identical structures could be extracted in the border regions. Both figures show combinations of the edge images. Edges in the ground truth image are combined with edges from the RC or MI registered image, respectively, by overlaying the edge images and color encoding the edges. In the combined edge images, red edges represent edges from the reference image, blue edges represent edges from the registered image and green edges represent overlapping edges from the reference and registered images. It can be seen that many of the extracted salient structures directly overlap and may thus be considered as correctly registered.

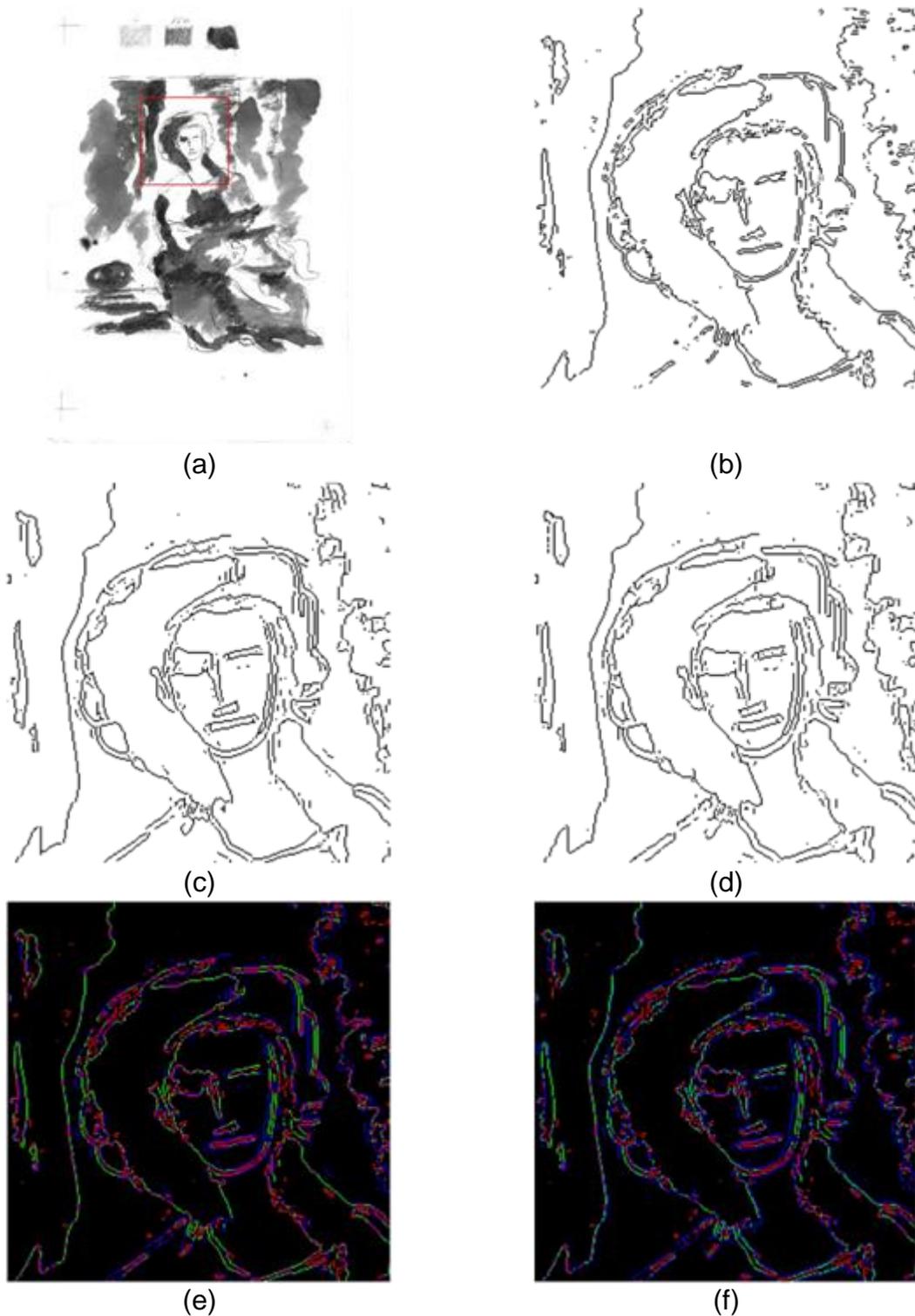

Figure 3.2: Edge images for the region with many details: (a) Reference image with extracted region; (b) edges in the reference image; (c) edges in the RC registered image; (d) edges in the MI registered image; (e) Combined edges from reference and RC registered image; (f) Combined edges from reference and MI registered image. In (e) and (f), red represents edges from the reference image, blue represents edges from the registered image and green represents overlapping edges from both images.

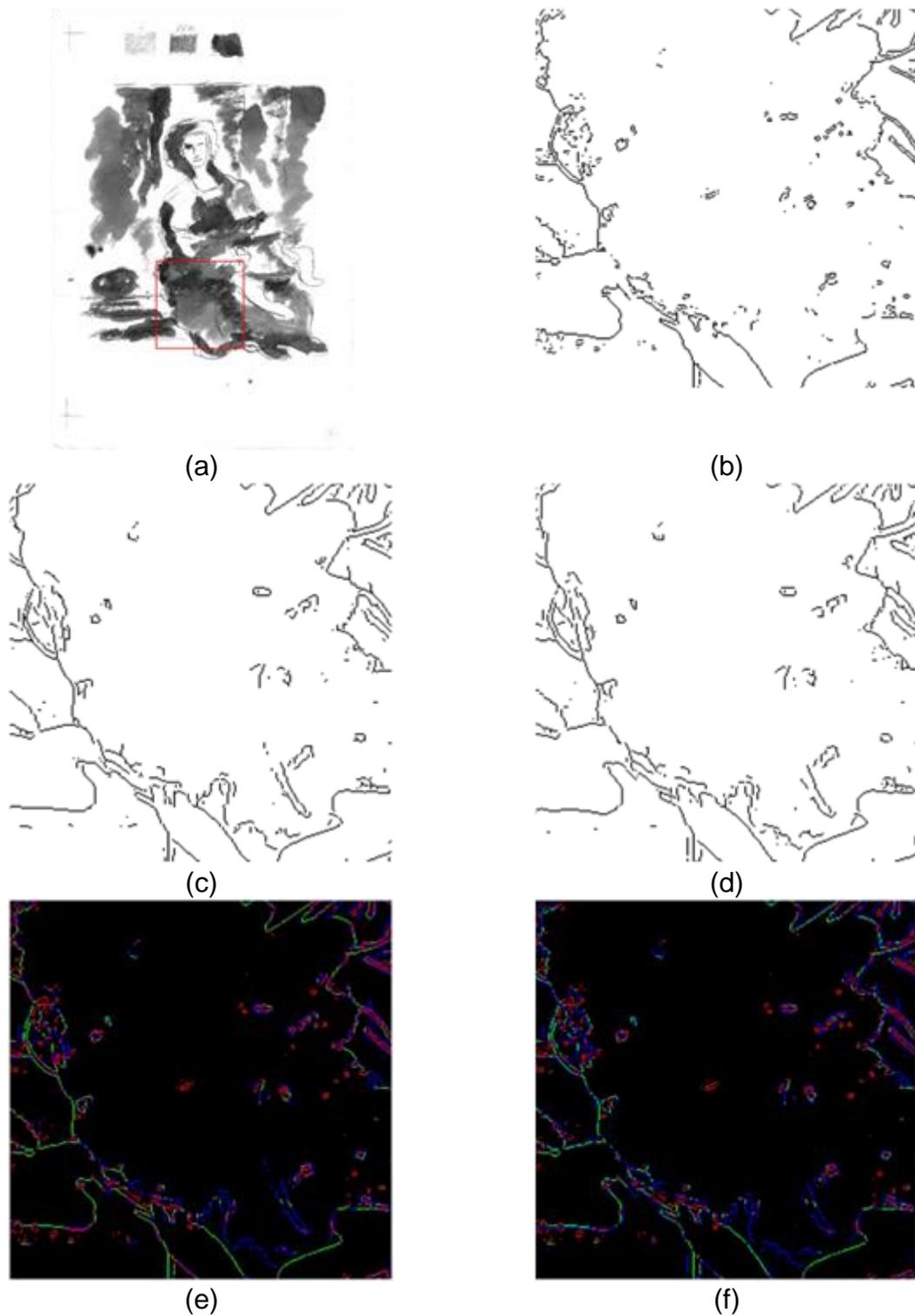

Figure 3.3: Edge images for the region with few details: (a) Reference image with extracted region, (b) edges in the reference image, (c) edges in the RC registered image, (d) edges in the MI registered image, (e) Combined edges from reference and RC registered image, (f) Combined edges from reference and MI registered image. In (e) and (f), red represents edges from the reference image, blue represents edges from the registered image and green represents overlapping edges from both images.

## 4 Discussion and Conclusion

In this work, we conducted a survey on different registration algorithms and investigated their suitability for hyperspectral historical image registration applications.

After the evaluation of different algorithms, we choose an intensity based registration algorithm with a curved transformation model. For the transformation model, we select cubic B-splines since they should be capable to cope with all non-rigid deformations in our hyperspectral images. From a number of similarity measures, we found that residual complexity and localized mutual information are well suited for the task at hand. In our evaluation, both measures show an acceptable performance in handling all difficulties, e.g., capture range, non-stationary and spatially varying intensity distortions or multi-modality that occur in our application.

As we showed in chapter 3, the registration using RC as similarity measure as well as the registration using the MI as similarity measure perform equally well in the region of interest of our application. In addition, both algorithms perform well in the two selected regions where one shows many and the other few details. Here, the method using RC performs slightly better in the high detailed region and the MI performs slightly better in the low detailed region. One possibility for this behavior might be that the RC uses a frequency encoding of the images to measure their similarity and therefore benefits from a higher amount of details, i.e., frequencies, in the images. Therefore, it might be advantageous to apply a registration using the RC as similarity measures for images that show a high amount of details and the MI otherwise.

**References**

[Dav17]     Davari, A.; Häberle, A.; Christlein, V.; Maier, A.; Riess, C.: "Sketch Layer Separation in Multi-Spectral Historical Document Images." Pattern Recognition Lab, Friedrich-Alexander University, Tech. Rep. (2017).

[Dic45]     Dice, Lee R. "Measures of the amount of ecologic association between species." Ecology, 3rd ser., 26 (1945): 297-302.

[Mid16b]    Middleton Spectral Vision. What are the key components of a hyperspectral system? July 2016. http://www.middletonspectral.com/what-are-the-key-components-of-a-hyperspectral-system/.

[Mid16c]    Middleton Spectral Vision. What is hyperspectral imaging? Accessed July 2016. http:// www.middletonspectral.com/what-is-hyperspectral-imaging/.

[Mid16a]    Middleton Spectral Vision. VNIR Cameras. Accessed July 2016. http://www.middletonspectral.com/products/hyperspectral-components-systems/hyperspectral-cameras/vnir-cameras/.

[Myr10]    Myronenko, A.; Song, X.: Intensity-Based Image Registration by Minimizing Residual Complexity, IEEE Transactions on Medical Imaging, Bd. 29, Nr. 11, Nov 2010, S. 1882–1891.

[Pre07]    Press, W. H.: Numerical recipes 3rd edition: The art of scientific computing, Cambridge university press, 2007.

[Sob68]    Sobel, Irwin and Feldman, Gary. "A 3x3 isotropic gradient operator for image processing." A talk at the Stanford Artificial Project in, 1968, 271-72.

[Thi98]    Thirion, J-P. "Image matching as a diffusion process: an analogy with Maxwell's demons." Medical image analysis 2, no. 3 (1998): 243-260.

**Appendix**

Numerous researchers have worked on registration algorithms and it is still an open topic of research. Consequently, there is a wide variety of different registration methods that are appropriate to use, dependent on the context. In this appendix, we provide additional details on the choice of the various choices that were made for selecting an algorithm and we provide additional results on the phantom dataset.

## 1.1 Introduction to Image Registration

When we capture multiple images from a drawing with one or more cameras, it is most unlikely that the captured images have an accurate pixel-wise alignment. Each lens may introduce individual geometric distortions or chromatic aberrations, camera sensors may exhibit imperfections, and the object itself may be in a slightly different pose for different acquisitions. Therefore, if after applying some image analysis, we need to compare the image with the original drawing/document, we need to find a geometrical transformation to compensate for the distortion introduced to the image. The process of finding a geometrical transformation, applying that on the image and make the captured image and the original image pixel-wise aligned is called image registration.

In 2D Euclidean plane, any point can be addressed via two elements; $x$ and $y$. Any 2D shape or image consists of many 2D points. Therefore, when applying a geometrical transformation on an image, we, basically, are applying that transformation to all the image points.

Mathematically, a 2D point in Euclidean plane can be shown as $\begin{bmatrix} x \\ y \end{bmatrix}$. The same 2D point can be represented as $p = \begin{bmatrix} x \\ y \\ 1 \end{bmatrix}$ in homogeneous coordinate by adding a third dimension (Fig. 1).

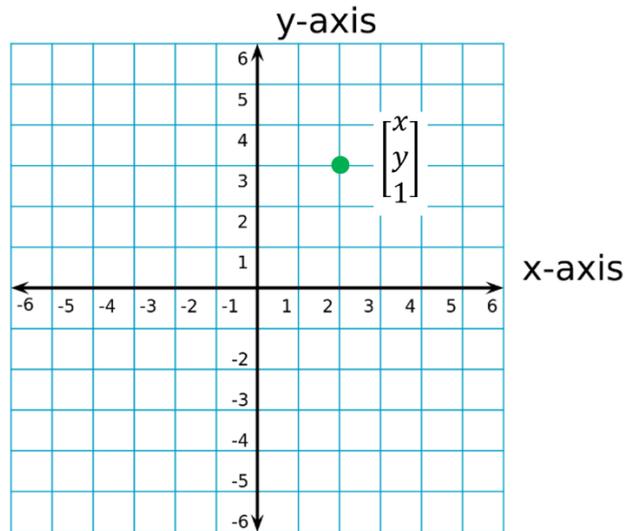
Figure 1: Representation of a 2D point in Euclidean plane using homogeneous coordinate.

A geometrical transformation $T$ in homogeneous coordinate however, is represented by a $3 \times 3$ matrix $T = \begin{bmatrix} a & b & c \\ d & e & f \\ g & h & i \end{bmatrix}$. Transforming the point $p$ to $p'$ using the transformation matrix $T$ is nothing but multiplying $T$ and $p$.

$$p' = Tp = \begin{bmatrix} a & b & c \\ d & e & f \\ g & h & i \end{bmatrix} \begin{bmatrix} x \\ y \\ 1 \end{bmatrix} = \begin{bmatrix} ax + by + c \\ dx + ey + f \\ gx + hy + i \end{bmatrix} = \begin{bmatrix} \frac{ax + by + c}{gx + hy + i} \\ \frac{dx + ey + f}{gx + hy + i} \\ 1 \end{bmatrix} \quad (2.1)$$

For instance, in Fig. 2 (a) and Fig. 2 (b) the green point is geometrically transformed and the new red point is generated. In Fig. 2 (a), the green dot is moved 2 units along the $x$ axis and $-3$ units along the $y$ axis. This geometrical transformation is called translation. On the other hand, the transformation in Fig. 2 (b) can be considered as a $\alpha - \beta$ degree rotation over the origin towards $x$ axis. This geometrical transformation is called rotation.

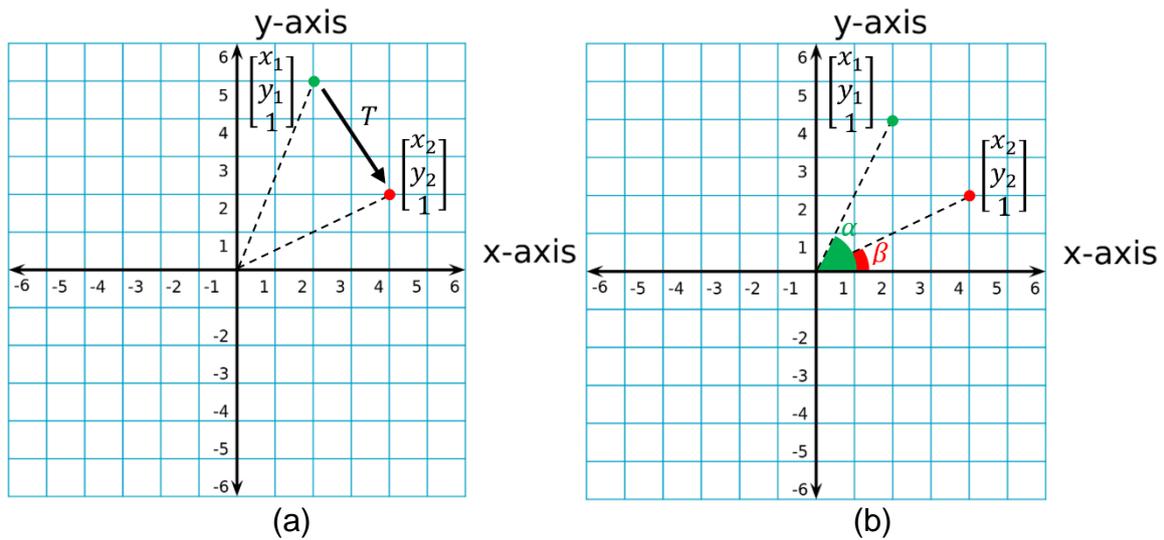

Figure 2: (a) Translation transformation, (b) Rotation transformation.

Considering more complex 2D shapes, of basic geometrical transformations we can name translation, rotation, scale and shear (Fig. 3 (a)-(e)).

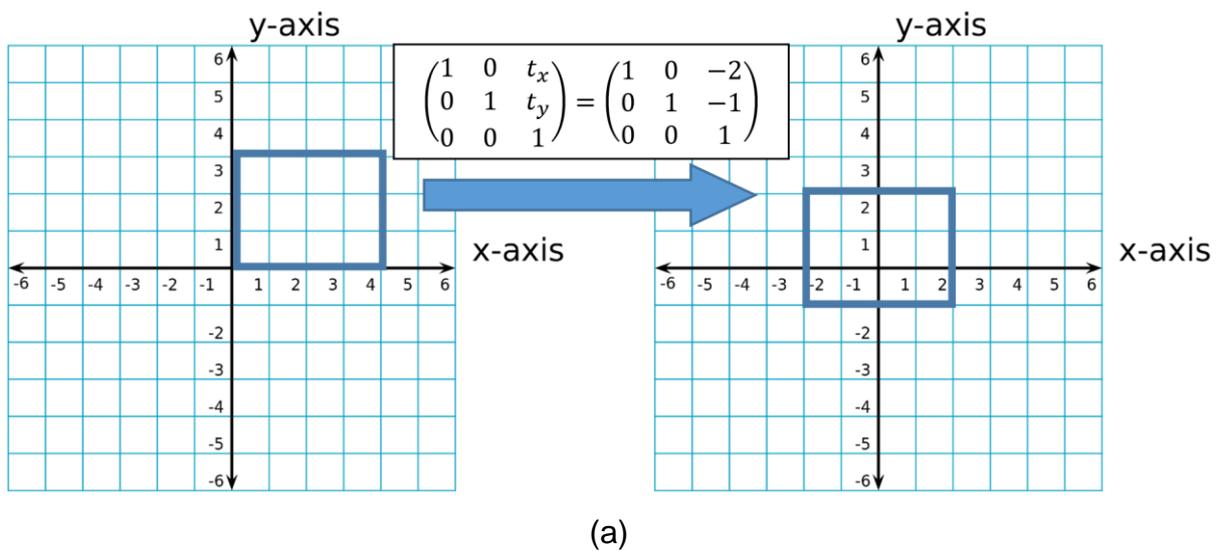

(a)

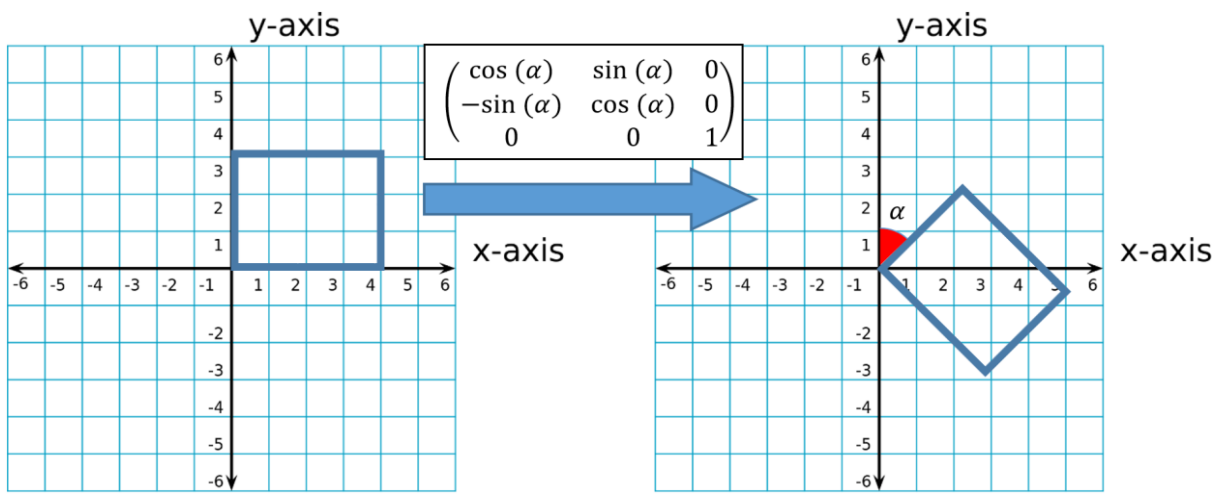

(b)

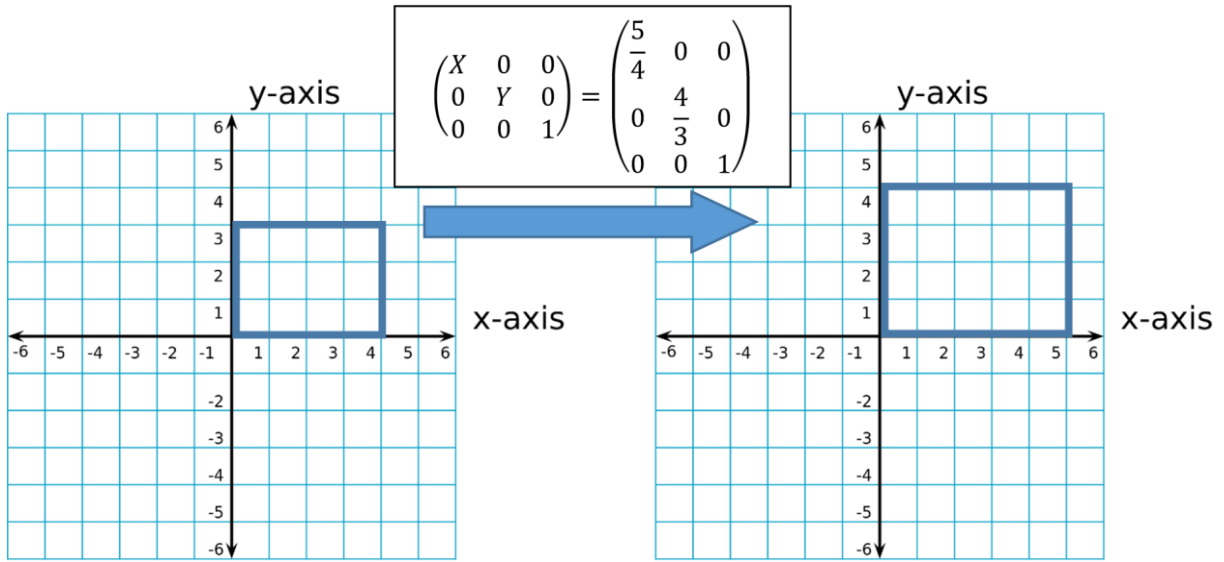

(c)

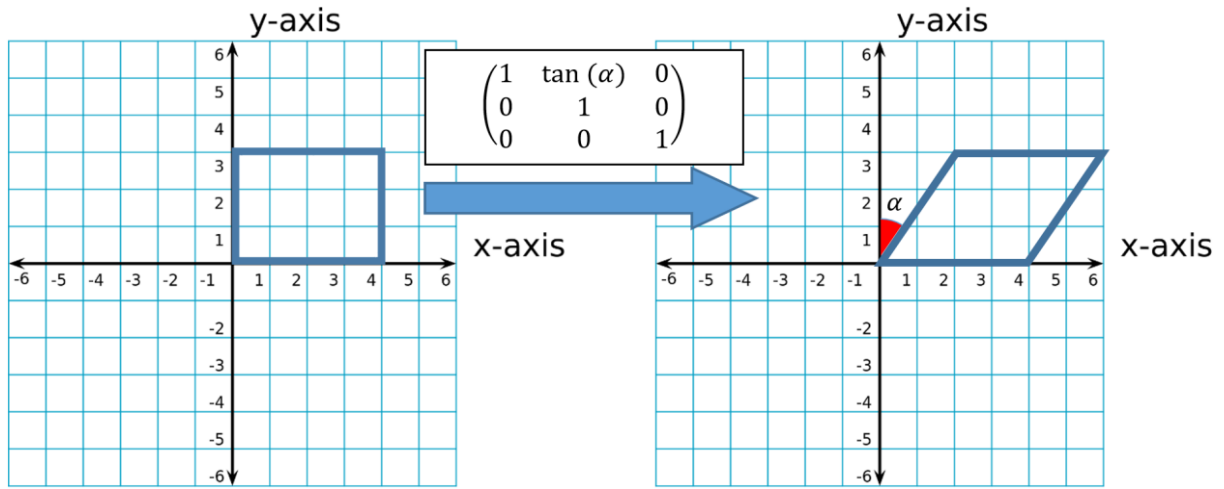

(d)

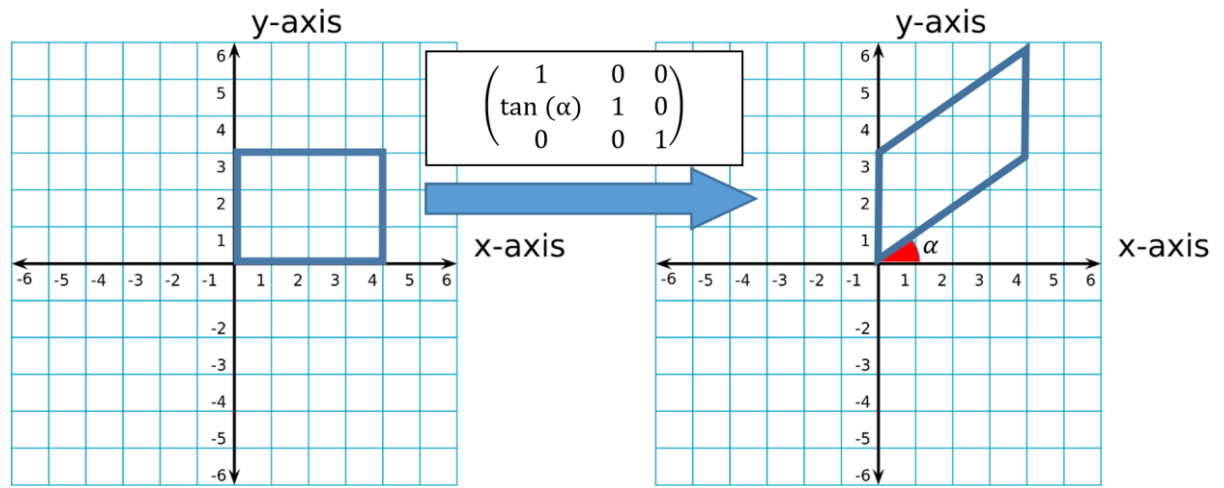

(e)

Figure 3: Instances of basic geometrical transformations with their transformation matrix $T$. (a) Translation, (b) Rotation, (c) Scale, (d) Shear along the $x$ axis, (e) Shear along the $y$ axis.

These basic geometrical transformations can combine together and create more complex transformations. Representing the transformations by matrices in homogeneous coordinate makes the combination if different transformations very easy. Fig. 4 (a) illustrates a transformation $T$ which is a combination of translation $T_T$ and rotation $T_R$. Therefore, the overall transformation is the multiplication of each transformation matrix

$$T = T_R \times T_T \qquad (2.2)$$

Fig. 4 (b) shows a transformation $T$ which is a combination of translation $T_T$, rotation $T_R$ and shear $T_S$. Same as in equation 2.2, the overall transformation matrix is generated by

$$T = T_S \times T_R \times T_T \qquad (2.3)$$

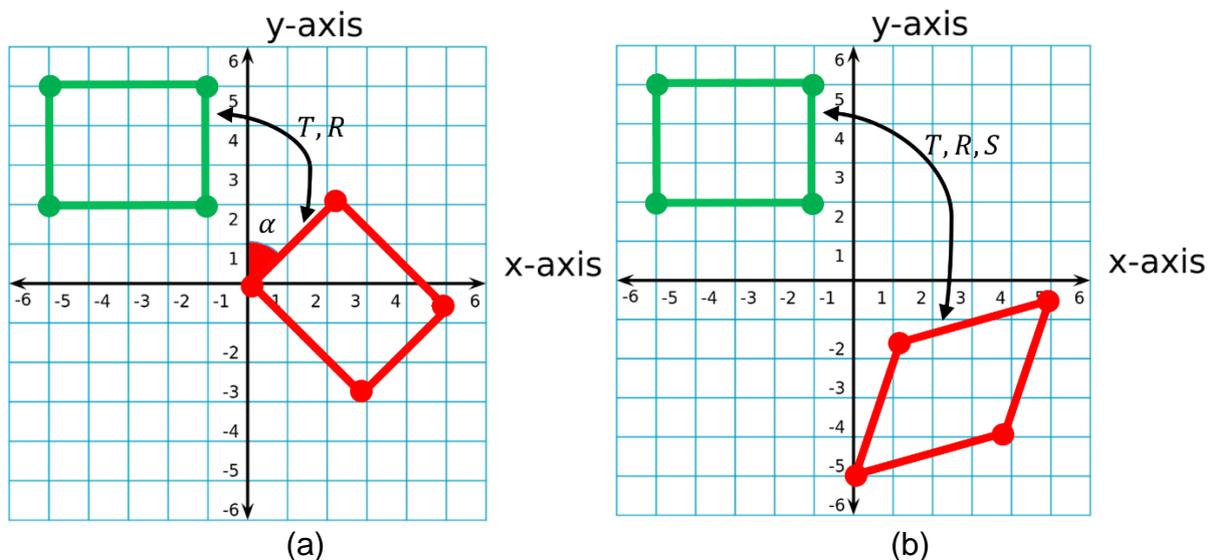

(a)          (b)

Figure 4: (a) a transformation which is a combination of translation $T_T$ and rotation $T_R$, (b) a transformation which is a combination of translation $T_T$, rotation $T_R$ and shear $T_S$.

The main task of a registration pipeline is to find/estimate the transformation matrix $T$. Not all transformations are as simple as the aforementioned ones. For instance, as shown in Fig. 5, no combination of the aforementioned transformations can produce the red trapezoid from the green rectangle. The reason is that the parallel lines in the green rectangle are not preserved in the red trapezoid. Therefore, we need a set of more complex transformations in order to be able to transform the green rectangle to the red trapezoid.

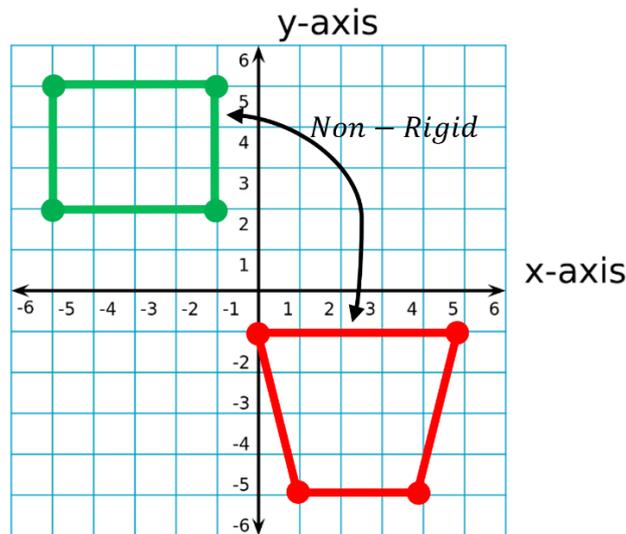
Figure 5: An example of a non-rigid transformation.

Any transformation, which preserves the points, lines, length, angles and the parallelism of a set of two lines, is called rigid transformation. Translation, rotation, reflection and any combination of them fall into the rigid transformation category. Any transformation, which preserves the points, lines and the parallelism of a set of two lines, is called affine transformation. Translation, rotation, reflection, shear and any combination of them fall into the affine transformation category. Any transformation that fall out of rigid and affine transformation categories is called non-rigid transformation.

The main goal of each image registration is to align images of the same, or in some cases, similar objects. These images are normally taken from different views and sometimes also using different modalities. Therefore, all registration methods search for a geometrical transformation that aligns points in one image as good as possible with the corresponding points in another image. However, registration is not limited to 2D image data. There also exist many applications for 3D or higher dimensional registration [Mar12] [Oli14].

To find such a transformation, the registration algorithm needs the images to be registered as input. In this work, we will refer to these images as the 'reference' image and the 'moving' image. The reference image may be seen as some kind of ground truth image while the moving image may be seen as a distorted copy of that image. Therefore, the moving image is registered onto the reference image, i.e., the reference image is kept fix while the moving image is transformed using a geometrical transformation. The transformed moving image is then evaluated against the reference image and the registration process is stopped if a previously specified criterion is met. The output of the algorithm is the geometrical transformation that registers the moving to the reference image.

Every registration algorithm consists of three main components; the transformation model, the similarity measure, and the optimization method [Cru14]. The transformation model specifies all changes that may be applied to the source image in the attempt to match it to the reference image. The similarity measure is a function that evaluates the difference between the reference and the transformed moving image. The difference is either computed as the overall distance between salient structures in the reference and moving image or by comparing pixel intensity distributions. To find an optimal registration, the similarity measure has to be optimized. This is done by the optimization method that tries to optimize the similarity measure by searching optima in the parameter space of the transformation model. Using these basic components, a mathematical description of a registration method may be given as

$$T = \max_{T} S(I, T(J)),\qquad(2.4)$$

where $T$ is the transformation model that registers the source image $J$ onto the reference image $I$ and $S$ is the similarity measure the optimization method has to find an optimum for.

## 1.2 Feature-Based vs. Intensity-Based Registration

Besides the three main components mentioned in section 1.1, another important aspect is whether the registration algorithm is feature-based or intensity-based. Generally, feature-based methods rely on geometrical structures that have to be extracted from both images, whereas intensity based methods exclusively rely on the voxel intensities [Oli14] [Roc99].

### 1.2.1 Feature-Based Registration

Feature-based registration methods are methods that are based on the segmentation of salient structures in the reference and moving image. These features may be distinct landmarks like significant regions, lines, points or surfaces. Furthermore, they have to be distributed over the whole image space and extractable from both images. Besides, their number has to be sufficiently high, independent of changes to image geometry, noise and changes in the scene [Zit03]. After feature extraction, an optimal registration is searched by optimizing the similarity measure. It tries to find an optimal solution for the matching of corresponding points by altering the parameters of the transformation method. For example, the similarity measure might search for a minimum of the overall distance of a set of point features by varying the parameter of the transformation $T$.

Putting feature-based methods into the context of the mathematical description of (2.1), the similarity measure $S$ and transformation $T$ are not applied to the whole images during optimization but only to the extracted features. After the optimization, the found transformation is applied to the source image. In the case of a rigid or affine transformation, which will be discussed in 2.3, this can be done directly by applying the transformation to every pixel. For other transformation models, the transformation has to be interpolated between feature points where it was not explicitly calculated [Fit00][Mai98].

One of the main drawbacks of the feature-based registration methods is that they contain an additional segmentation step for the feature extraction that may not be trivial and will set a limit to the accuracy of the subsequent registration. It might be especially hard to find enough features of sufficient quality over the whole image space under the presence of noise or under the use of different sensor types [Mai98][Zit03].

### 1.2.2 Intensity-Based Registration

In contrast to the feature-based methods, the intensity-based methods work directly on the pixel intensities without trying to extract any features. They define similarity measures that exploit mathematical or statistical dependencies between the image intensity values. Using these measures, they search iteratively for the transformation that optimizes the similarity measure when it is applied to the moving image. Intensity-based methods expect the moving and reference image to be most similar at the correct alignment and can only be computed on the overlapping regions of the images. An example of an intensity-based similarity measure would be the sum of squared differences (SSD) [Fit00][Haj95]. The SSD is probably the easiest and most intuitive of the intensity-based methods. Irrespective of Gaussian noise, SSD expects the images to be identical when registered and therefore computes the difference between intensity values with respect to the L2-norm [Cru14][Oli14]. In the context of equation (2.4), this means that the similarity measure $S$ compares intensity values in the images $I$ and $T(J)$ to find the transformation.

One drawback of the intensity-based methods is their high computational complexity due to the evaluation of the whole image. In our application, more important drawback is that these methods tend to converge to a local optimum, especially in the presence of noise. However, the probability of converging to a local optimum can be reduced and the speed of convergence increased by using a pre-registration method. A pre-registration is an additional, simpler registration that is applied before a more complex one and reduces the initial error between moving and reference image [Oli14]. Another possibility to increase the convergence speed and reduce the risk of converging to a local optimum is to use a pyramidal approach.

The pyramidal approach is a multi-resolution strategy [Ade84]. The reference and moving image are down-sampled to generate two image pyramids. The registration is then computed by consecutively registering the down-sampled versions of the reference and moving image, starting at the lowest resolution [Oli14].

### 1.2.3 Selected Descriptor: Intensity-based

For the decision between feature-based and intensity-based registration, the amount of noise and blur in the images is of high interest. Furthermore, different sensors were used for the acquisition of the ground truth and spectral data. As it can be seen in Fig. 6, the spectral images are affected by a relatively high amount of noise and show heavily blurred edges, dependent on the spectral channel. Especially the heavy blur of the edges will make it hard to find enough features with sufficient quality in the spectral images and using features with insufficient quality will highly degrade the registration result. The intensity-based methods will be affected by the noise and blur in the spectral images as well since they will lead to a higher number of local optima in the optimization. However, using a pre-registration algorithm and pyramidal scheme in the optimization approach will reduce the probability of getting stuck in a local optimum. Therefore, we conclude that an intensity-based registration algorithm is best suited to register the spectral and ground truth images.

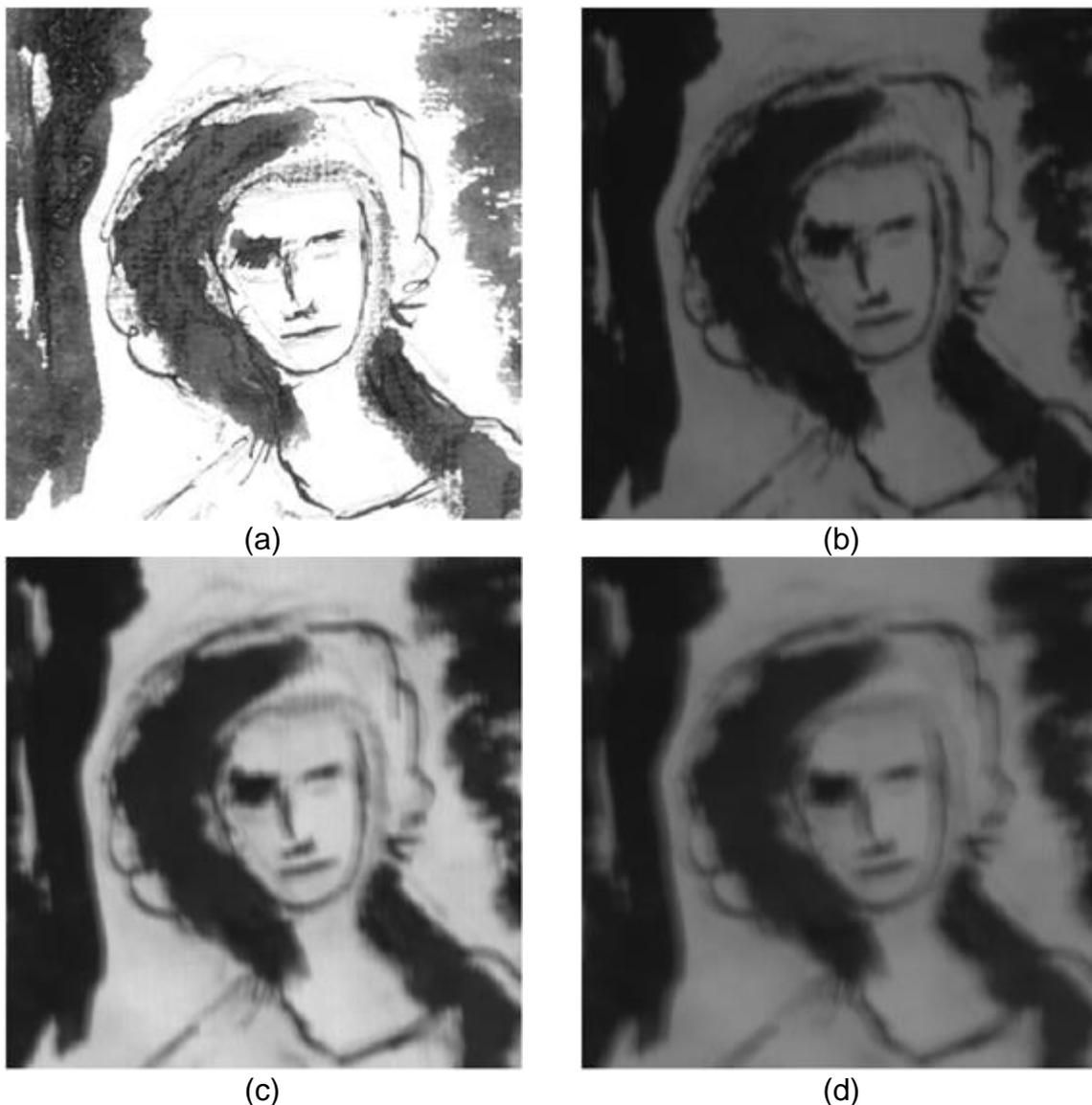

Figure 6: Presence of noise and blur in the spectral channels of a hyperspectral image: (a) ground truth, (b) channel 74, (c) channel 104, (d) channel 144; The spectral channels are from a sample hyperspectral image of phantom data containing 258 spectral channels.

## 1.3 Transformation Models

The selection of an appropriate transformation model is crucial to the success of every registration method. It defines all transformations that may be applied during image matching and has to be chosen dependent on the deformations in the input data [Cru14][Oli14]. In the literature, numerous models can be found that can usually be divided into rigid, affine and non-rigid transformations as it is illustrated in Fig. 7.

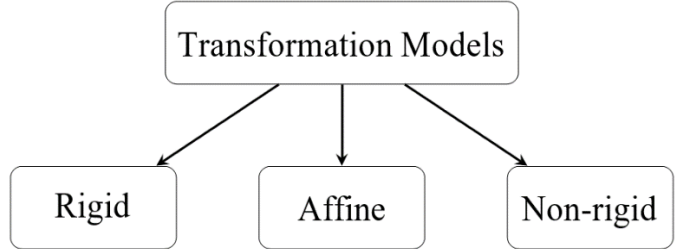

Figure 7: Overview of the different transformation models.

### 1.3.1 Rigid Transformations

The rigid transformations are probably the most intuitive and widest known transformations. They are restricted to preserve all distances between lines and thus allow only translations and rotations. In the three dimensional case, it can be defined using only six parameters, three translations and three rotations.

In the two dimensional case, the number is reduced to only three, two for the translation and one for the rotation [Gol65] [Har03]. Using a matrix and vector notation with homogeneous coordinates, a rigid transformation in 2-D can be specified as:

$$\begin{bmatrix} j'_x \\ j'_y \\ 1 \end{bmatrix} = \begin{bmatrix} R & t \\ \hline 0 \quad 0 & 1 \end{bmatrix} \begin{bmatrix} j_x \\ j_y \\ 1 \end{bmatrix} \qquad (2.5)$$

where $j_x$ and $j_y$ are the $x$ and $y$ coordinates of a point in the source image $J$ and $j'_x$ and $j'_y$ are the coordinates after transformation. $R$ is a two dimensional rotation matrix and $t$ a translation vector. They are specified as:

$$R = \begin{bmatrix} \cos(\alpha) & -\sin(\alpha) \\ \sin(\alpha) & \cos(\alpha) \end{bmatrix}, \quad t = \begin{bmatrix} t_x \\ t_y \end{bmatrix} \qquad (2.6)$$

where $\alpha$ is the angle of the rotation in the $x$-$y$ plane and $t_x$ and $t_y$ define the translation in $x$ and $y$ direction. Allowing only rotations and translations, rigid translations are highly restricted. Thus, in many cases they are unable to find a proper registration. For example, this may be the case with deformed objects or when the scene is non-rigidly distorted. Nevertheless, they may be used in a pre-registration [Fit00][Oli14].

### 1.3.2 Affine Transformations

Another type of transformations that may be applied during image registration are the affine transformations. They are restricted to preserve collinearity and allow besides rotation and translation also scaling and shearing. This leads to six additional parameters in the definition of the three dimensional case, three for scaling and three for shearing and four additional parameters in the two dimensional case, two for scaling and two for shearing [Gol65] [Har03]. Using a matrix and vector representation with homogeneous coordinates, an affine registration in 2-D can be represented similar to equation (2.5) as:

$$\begin{bmatrix} j'_x \\ j'_y \\ 1 \end{bmatrix} = \begin{bmatrix} A & t \\ 0 \quad 0 & 1 \end{bmatrix} \begin{bmatrix} j_x \\ j_y \\ 1 \end{bmatrix} \qquad (2.7)$$

However, where $R$ contains only the rotation parameters in the rigid case, A also contains the parameters for scaling and shearing in affine transformations [Fit00]. Although, they are less restrictive than rigid transformations, affine transformations are still unable to cope with more free deformations like those that may be caused by deformed objects or lens distortions.

### 1.3.3 Non-rigid Transformations

The curved transformations are, at least in medical applications where transformations are often non-rigid, the most important and widest used transformation types. Generally, they can be divided into free-form transformations and guided or elastic transformations. Where free-form deformations allow basically any deformation, guided transformations are regulated by a physical model that is derived from the underlying materials in the images [Cru14] [Oli14].

The most popular free-form deformation method is the cubic B-spline. The method defines a grid of control points that are moved individually to optimize the similarity measure. Splines between the control points are then only defined in the neighborhood of the control points and thus have only a local effect on the transformation [Cru14][Oli14]. The method was originally introduced to the registration community by

Rueckert et al. [Rue99]. The local B-spline transformation for the three dimensional case $T$ is given as:

$$T(x, y, z) = \sum_{l=0}^{3} \sum_{m=0}^{3} \sum_{n=0}^{3} B_l(u) B_m(v) B_n(w) \phi_{i+l, j+m, k+n}, \quad (2.8)$$

where $\phi$ is a $n_x, n_y, n_z$ mesh of control points with uniform and $x, y, z$ defining positions in the image volume. Further, $i = \lfloor \frac{x}{n_x} \rfloor - 1, j = \lfloor \frac{y}{n_y} \rfloor - 1, k = \lfloor \frac{z}{n_z} \rfloor - 1, u = \frac{x}{n_x} - \lfloor \frac{x}{n_x} \rfloor, v = \frac{y}{n_y} - \lfloor \frac{y}{n_y} \rfloor$ and $w = \frac{z}{n_z} - \lfloor \frac{z}{n_z} \rfloor$. The $B_l$, $B_m$ and $B_n$ represent the basis functions of the B-spline:

$$B_0(u) = \frac{(1-u)^3}{6}, \quad B_1(u) = \frac{3u^3 - 6u^2 + 4}{6},$$

$$B_2(u) = \frac{-3u^3 + 3u^2 + 3u + 1}{6}, \quad B_3(u) = \frac{u^3}{6}$$

The degree of non-rigid deformation that can be modeled is highly dependent on the control point mesh resolution. Global deformations can be modeled using a low density of grid points, while local deformations are modeled using a high density. Further, the degree of freedom is dependent on the number of control points and thus the computational complexity [Rue99]. Shortly after the method was introduced, it was extended to deformable grids by Schnabel et al. [Sch01]. The method's main advantage is its high flexibility while still being computationally efficient, what has led to a wide use of the method in many applications [Cru14].

The guided or elastic transformations use physical models based on the material properties of the image content to estimate a deformation field for the registration. They were introduced by Bajcsy et al. [Baj89] where brain images are modeled as pieces of rubber and deformed in a coarse-to-fine strategy. To obtain a registration using these methods, two types of forces are modeled that effect the structure of the elastic solid. The first type are external forces that are derived from the similarity measure and try to force the image toward an optimum of the measure. The second type are internal forces. They are dependent on the modeled physical properties and oppose the deformation by the external forces. A registration is then accomplished by stretching and compressing the image until the external and internal forces reach a state of equilibrium [Oli14] [Zit03].

### 1.3.4 Selected Transformation: Cubic B-spline Non-rigid Transformation

For the choice of a valid transformation model that is capable of registering the source and target images, the deformations between the images are the crucial criterion. Possible sources of such deformations may be different positioning of the sensors or the object, different sensors and lenses or the deformations of objects in the scene between image acquisitions.

In our case, deformations of objects between image acquisition is not an issue, simply because there are not going to be any deformable objects in historical images. As it can be seen in Fig. 8, the lens of the spectral camera introduces distortions to the images, e.g. barrel distortions. These distortions are non-rigid and are therefore not remediable by rigid transformations. Further, affine transformations are insufficient too since they preserve for example the straightness of lines. Consequently, the chosen transformation will be a curved one. From the variety of curved transformations, we choose the cubic B-splines due to their wide applicability and ability to cope with local and global deformations.

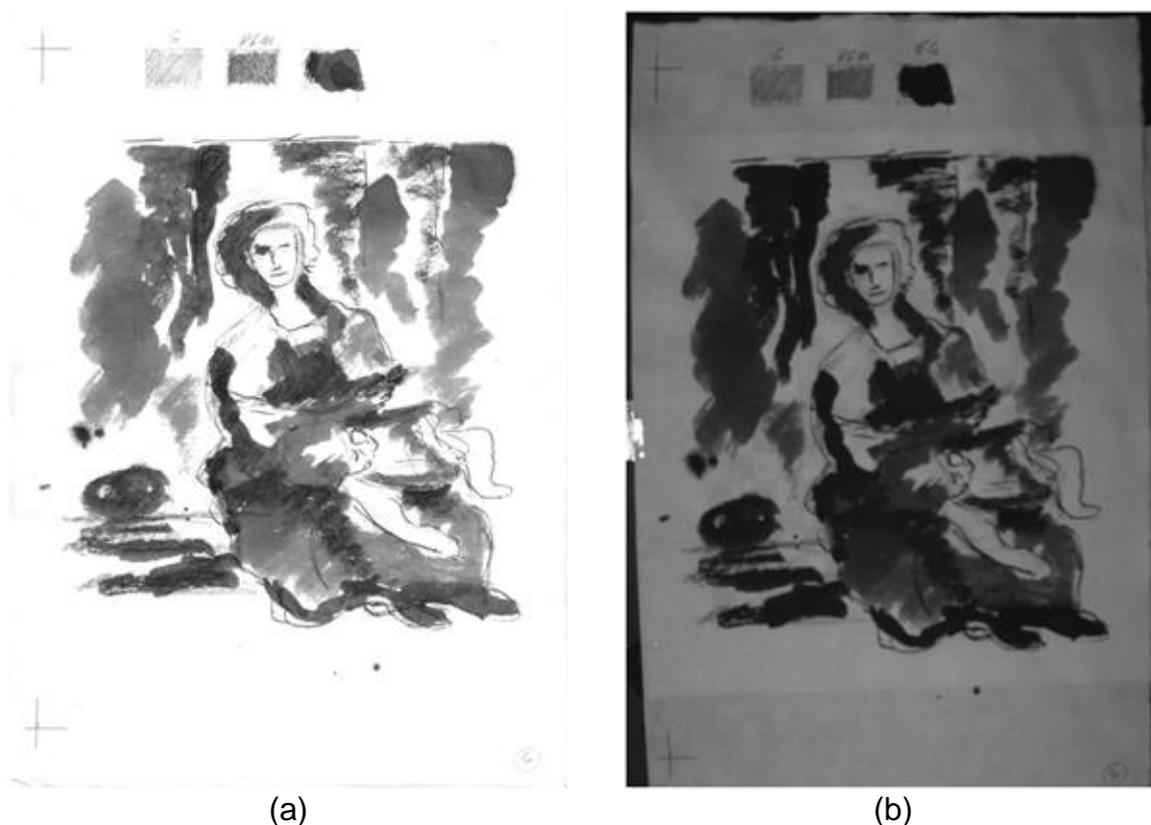

(a) (b)

Figure 8: (a) The reference image and (b) channel 100 of the 258-spectral channels image recorded. Barrel distortion can be seen at the left edge of the image in the spectral channel. For an example of the uneven surface of paper and ink see the top right corner of the spectral channel. Over the whole image space, pixel intensities appear darker in the spectral image.

Deformations due to different sensor positioning may or may not occur dependent on whether different sensors are used at all and if so, on the positioning accuracy. For our data, the ground truth and hyperspectral data were acquired with two different sensors and it appears that the acquisition geometries were not perfectly equivalent. This results in a rigid deformation of the images. Therefore, we apply a rigid pre-registration before using a free-form transformation.

## 1.4 Intensity-Based Similarity Measures

The success of the registration is also highly dependent on the choice of a proper similarity measure. There exist several different intensity-based similarity measures and also variations of these measures. Most of these measures are either based on intensity differences, cross- correlation of intensity values or information theory [Oli14].

### 1.4.1 Sum of Squared Distances (SSD)

As it was mentioned in section 1.2.2, SSD is probably the most intuitive intensity-based similarity measure and optimal under the assumption that the images differ only due to a misalignment and Gaussian noise. For source image $J$ and reference image, the measure is computed as:

$$\text{SSD} = \frac{1}{N} \sum_{i=1}^{N} |I_i - T(J_i)|^2, \qquad (2.9)$$

where $i$ is a pixel in the compared images and $N$ the number of pixels in the overlapping area. The method may be optimal under the assumption of only Gaussian noise and misalignment, but it will fail if the images divert too far from these assumptions. This is, for example, the case if some corresponding pixels have high intensity differences or for non-Gaussian noise [Fit00].

### 1.4.2 Cross Correlation (CC)

Another type of similarity measures that may be used in registration is the cross correlation (CC) and its related methods like the Pearson's correlation coefficient and the correlation ratio [Oli14].

The cross correlation and correlation coefficient assume a linear relationship between the pixel intensity values in the reference and source image. The correlation coefficient is computed using the standard deviations $\sigma_I$ and $\sigma_{T(J)}$ and mean values $\mu_I$ and $\mu_{T(J)}$ of the reference image $I$ and transformed moving image $T(J)$ as:

$$\text{CC} = \frac{\sum_{i=1}^{N}(I_i - \mu_I)(T(J_i) - \mu_{T(J)})}{\sigma_I \sigma_{T(J)}}, \qquad (2.10)$$

Where $I_i$ and $J_i$ denote the $i$-th pixel in image $I$ and $J$ respectively. These methods are applicable if the reference and moving image are acquired using the same sensor type. Besides, they assume image intensities to be independent and stationary from pixel to pixel and thus, are not applicable if these conditions are not met [Oli14].

The correlation ratio was introduced by Roche et al. as a similarity measure [Roc98]. It assumes a functional dependency between the images and produces values from zero to one, with one representing a purely deterministic dependence. Using the mean and variance of the overlapping image regions of the reference and moving image, it can be computed as:

$$\eta(I,J) = 1 - \frac{1}{N\sigma^2}\sum_{k} N_k \sigma_k^2, \qquad (2.11)$$

where $\sigma$ is the total and $\sigma_k$ the conditional variance and $N_k$ the number of pixels in the iso-set $k$. The correlation ratio is capable of registering images from different sensor types but like the CC, assumes image intensities to be independent and spatially stationary for corresponding pixels.

### 1.4.3 Mutual Information (MI)

The mutual information (MI) similarity measure was introduced simultaneously by Collignon et al. [Col95] and Viola et al. [Vio97] and, similar to the correlation ratio, assumes a functional dependency between intensity values in the reference and moving image [Oli14]. It is based on the image entropies $H(I)$ and $H(T(J))$ and their joint entropy $H(I,T(J))$ [Sha01] and tries to maximize the information that is shared by both images as:

$$\text{MI}(I,T(J)) = H(I) + H(T(J)) - H(I,T(J)) = \sum_{i \in I}^{N}\sum_{j \in J}^{N} p(i,j) \log \frac{p(i,j)}{p(i)p(j)}, \qquad (2.12)$$

where $p(i)$ and $p(j)$ are the marginal and $p(i,j)$ is the joint probability distribution of the image intensity values $i$ and $j$ in the overlapping area of $I$ and $T(J)$. This can be interpreted as so that the measure tries to eliminate the dispersion in the joint histogram of $I$ and $J$, caused by misregistration, by maximizing the interdependency of the images intensity values. It tries to make the joint histogram as non-uniformly distributed as possible.

The mutual information is suitable for the registration of data from different sensor types but is dependent on the overlap region of the images. To make the measure invariant of the overlap region, Studholme et al. [Stu99] introduced the idea of normalized mutual information (NMI). To accomplish this, the measure has to be made independent of changes to the image entropies $H(I)$ and $H(T(J))$. This is done by calculating the ratio between $H(I)$, $H(T(J))$ and the joint entropy $H(I, T(J))$ as:

$$\text{NMI}(I, T(J)) = \frac{H(I) + H(T(J))}{H(I, T(J))}. \tag{2.13}$$

As this makes the measure invariant to changes to the overlapping region, it still assumes image intensities to be independent and spatially stationary. One possible approach to overcome these assumptions is to localize the similarity measure as it was done for example by Studholme et al. [Stu06] or Klein et al. [Kle08]. The main idea behind these methods is the assumption that although image intensities may not be independent and spatially stationary over the whole image, they may be assumed to be, within small neighborhoods.

The approach by Klein et al. [Kle08] computes the mutual information on several subregions of the image and then generates the localized mutual information (LMI) by averaging the mutual information from the sub-regions as:

$$\text{LMI}(I, T(J); \Omega) = \frac{1}{N} \sum_{x_j \in \Omega} \text{MI}(I, T(J); N(x_j)), \tag{2.15}$$

where, $\Omega$ represents the image domain and $N(x_j)$ a spatial neighborhood within the domain, centered on the coordinate $x_j$. $N$ is the number of neighborhoods.

The localized mutual information methods are capable of dealing with spatially-varying intensity distortions and are therefore often used for the registration of magnetic resonance tomography images (MRI) [Myr10].

### 1.4.4 Residual Complexity (RC)

The residual complexity (RC) is, like the sum of squared distances, also based on the pixel intensity differences, i.e., residual, between the reference and moving images. It was introduced by Myronenko et al. [Myr10] as a similarity measure that is able to deal with spatially varying intensity distortions. After the computation of the residual between reference and moving images, the RC uses the basis functions of the discrete cosine transform (DCT) to encode the residual. For a good registration result the complexity of the residual should be as low as possible, which is expressed by a high number of zero coefficients in the DCT. Therefore, the RC prefers a result with a high number of zero coefficients and few high coefficients over one with many small coefficients. In contrast to SSD, the RC does not favor an all zero residual, but one that can be sparsely coded with only a small number of basis functions and is computed as:

$$E(T) = \sum_{m=1}^{M} \log \frac{(\mathbf{q}_n^T \mathbf{r})^2}{\alpha + 1}, \qquad (2.16)$$

where $\mathbf{q}_n^T$ represents a basis function of the discrete cosine transform and $\alpha$ is a trade-off parameter coming from a regularization with the Kullback-Leibler divergence [Kul51]. The residual vector $\mathbf{r}$ is computed from the reference and moving image as:

$$\mathbf{r} = I - T(J) \qquad (2.17)$$

The main disadvantage of the residual complexity is that it relies directly on the difference of the images. This makes it not applicable if the intensity values in the input images vary much like it might occur in applications using different sensors or different exposure parameters.

### 1.4.5 Selected Similarity Measure: Mutual Information or Residual Complexity

The selection of a suitable similarity measure is highly dependent on the input data of the registration process. In our case, one important criterion for the measure is the capture range. The capture range describes how big the misalignment between the reference and moving images may be so that the similarity measure is still capable of finding the correct registration. Further, the measure has to be able to deal with non-stationary and spatially-varying intensity distortions and be able to cope with multi-modality sensor data.

The problem of the capture range can be overcome by using a pre-registration before applying the more precise registration so that all measures appear applicable under this condition. The second problem however is harder to overcome. The non-stationary and spatially-varying intensity distortions in our data exclude basically all presented measures except for the regional mutual information and the residual complexity. The residual complexity appears to be superior to the other methods under non-stationary and spatially-varying intensity distortions [Myr10]. However, the RC was designed for the registration of mono modal images while we are registering images from different sensor types. Concerning this, we think that the method will be applicable in our case anyway since the range of intensity values in the region of interest is comparable. There are no cases where the intensity values differ completely as it can be seen in Fig. 8. Consequently, we apply both the residual complexity and the localized mutual information methods to our data and compare the results in order to find the method that fits best to our application.

## 1.5 Optimization Method

In most cases, the task of the optimization method is to optimize the similarity measure by finding an optimal set of parameters for the transformation as it is the case for the cubic B-splines. Generally, every optimization may be referred to as a minimization. Instead of maximizing a function, one might just as well minimize its inverse. This leads to the following formulation of (2.14) for the optimization of the similarity measure:

$$T = \min_T S(I, T(J)), \qquad (2.18)$$

where $T$ is the transformation model that registers the source image $J$ onto the reference image $I$ and $S$ is the similarity measure. The method has to find the global minimum, which is supposed to represent the true alignment of the images, reliably and in an acceptable time frame. Finding the global minimum is normally not a trivial problem since the similarity measure often contains a multitude of local minima, a problem that grows bigger with the number of parameters in the optimization. This may be approached by applying a good pre-registration and a pyramidal scheme during the optimization since the pyramidal scheme will reduce the number of local optima and the pre-registration will shift the starting position for the optimization closer to the global optimum. However, there is no guaranty for convergence to the global minimum.

The time needed for the optimization is also highly dependent on the number of parameters to optimize. This is especially of concern for free-form deformations that allow for a high degree of flexibility since there are more parameters needed to express this flexibility.

We choose the steepest descent method [Pre07] for the minimization of our objective function as it was also used in [Myr10]. The steepest descent method is a line search method that are basically defined by their step length $\alpha_s$ and the direction of the search $p_s$ which is the negative gradient of the function in the steepest descent case at a step $s$ of the optimization. They compute a minimum iteratively where a step of the iteration process is given as:

$$x_{s+1} = x_s + \alpha_s p_s, \qquad (2.19)$$

where $x_s$ is the current position in the search space and $x_{s+1}$ is the position at the next step. The steepest descent method may be slow, dependent on the objective function. However, our application has no fix time constraints that have to be met and thus the convergence speed of the algorithm is of less concern. Furthermore, the optimization should start relatively close to the global minimum due to the applied preregistration.

## 1.6 Validation Metric

After the development of a registration algorithm, it is important to evaluate the proposed algorithm. However, as intensity-based similarity measures are not directly linked to registration errors in the data, the improvement of the error cannot be given directly by the measure. Further, there does not exist a standard procedure for the evaluation of intensity-based registration methods and the literature mentions a variety of different validation techniques.

The probably most obvious way to validate the registration result is to visually inspect it. Visual inspection is a fast and simple possibility to assess the general quality of the registration and is often used for roughly turning the registration parameters. However, it is inaccurate and highly subjective and additional methods are needed that yield quantitatively comparable results, especially for the fine-tuning of the registration parameter and for the comparison of different methods.

Another widely used approach is to apply an artificial transformation field to an image and thus create artificially transformed images. The images are then registered to the original image using the registration method that has to be evaluated. For the evaluation of the results it is especially beneficial that the deformation field, i.e., the artificial deformation, is known. Therefore, the transformation fields for deformation and registration may be directly compared. This method is often used to show that a registration method works in general but the artificial transformations often don't accord with the transformations occurring in the real world [D'A03][Myr10][Wan05].

The target registration error (TRE), i.e., the distance between corresponding points in the reference and registered moving image is an important measure for evaluating the quality of registration. The TRE is mainly used in point based registration and relies on fiducial markers that have to be extracted from the images. Its main disadvantage is its dependency on the fiducial localization error (FLE), i.e., the localization error of the fiducial markers. Furthermore, it only evaluates the registration at specific points and a conclusion about the registration quality in other areas is only possible if the registration in these areas is rigid [Dan11][Fit98].

Other methods try to evaluate the consistency [Chr01] and transitivity [Chr03] of the algorithm. To be consistent, the transformation from the moving to the reference image should be the same as the inverse transformation of the reference to the moving image and thus their combination should be as close to the identity mapping as possible. The transitivity evaluates the difference between the combination of the transformations between three images, i.e., A to B, B to C and C back to A with the identity mapping and should be as small as possible. The consistency and transparency of a registration algorithm may be evaluated using the cumulative inverse consistency error (CICE) and the cumulative transitivity error (CTE) [Chr06].

Another possibility to evaluate the registration result is the Dice similarity coefficient (DSC) [Dic45] or relative overlap (RO) [Chr06]. The DSC and RO measure the regional overlap of segmented areas in the reference and registered images as $\frac{area(P \cap S)}{area(P \cup S)}$ where $P$ is an area segmented from the reference image and $S$ the corresponding area segmented from the moving image. The measures return a value between 0 and 1, with 0 indicating no overlap and 1 indicating a complete overlap of the segmented areas. The DCS or RO were for example applied for evaluation in [Bha01], [Loe10] and [Ver07].

Further, there are several metrics that are based directly on the intensity values in the images. Such metrics include the root-mean square of intensity differences (RMSint), the median absolute deviation of intensity differences (MADint) or the maximum intensity difference (MAXint) [Urs07]. However, they are not very well suited for the evaluation of the similarity between multimodal images since intensity values between corresponding pixels may differ highly in multimodal applications.

### 1.6.1 Selected Metric: DSC Coefficient

For the selection of an appropriate validation metric, it is important to assure whether the selected metric provides a meaningful evaluation of the registration accuracy.

Visual inspection is applicable to all cases of image registration but is also very limited in terms of a meaningful evaluation and always has to be combined with other metrics. We use it for a general assessment of the applicability of the algorithms. The approaches based on artificial transformation fields are also applicable to all kinds of registration problems.

However, they are most often used to show that a registration strategy is working in general which has already been proven for the MI as well as the RC. Furthermore, non-rigid registration methods do not generally fulfill the consistency and transitivity properties [Cru14]. Therefore, the CICE and CTE metrics do not necessarily produce meaningful results for our application. In addition, the intensity-based metrics are not applicable in our case since they are not well suited for multi-modality applications. The TRE and DSC are in general both applicable for the evaluation of multi-modality registration. However, the TRE assumes the transformations between the segmented points to be rigid which conflicts with the B-spline transformations that we apply. Therefore, we will use DSC as the measure for the evaluation of the registration accuracy in our application.

## 2 Results

The hyperspectral images are registered to the reference image using the RC algorithm by Myronenko et al. in [Myr10] and the MI algorithm by Klein et al. [Kle08].

Both algorithms are part of the Medical Image Registration Toolbox (MIRT), implemented in Matlab, which can be downloaded at [Myr16]. A Flowchart of the performed registration can be seen in Fig. 9.

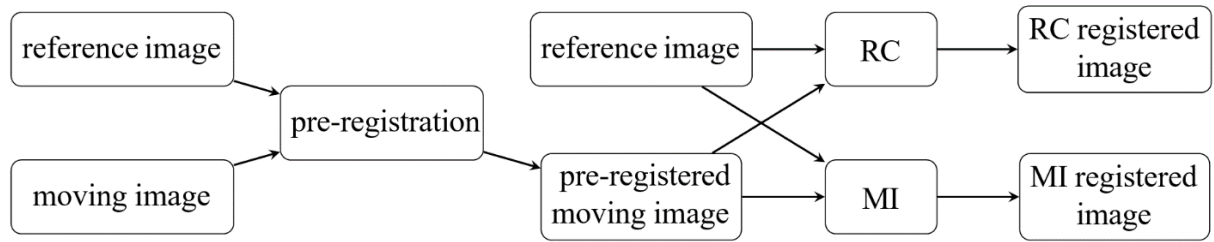

Figure 9: Flowchart for the complete registration procedure; first, a pre-registration is applied to the moving images; second, the pre-registered images are registered with the RC and MI algorithm respectively.

We evaluate the results of both proposed algorithms on the region of interest for our application, which is given in Fig. 10, and two additional regions from the moving images given in Fig. 11 and 12. To evaluate the algorithm under different circumstances we choose one region that shows a high amount of details, i.e., a high amount of high frequency components, and one region that shows few details, i.e., a low amount of high frequency components. For the evaluation, we suggest to first visually inspect the registration results. To make the visual inspection clearer, we also compute edge images of the two selected regions. The edges were extracted using the Sobel edge detector [Sob68]. Further, we compute the DSC to be able to numerically compare the registration results.

As it can be seen in Fig. 10, both algorithms perform very well in the central regions of the image, where it shows many structures. However, the registration fails in the border regions, if artifacts like the black area at the bottom of the moving images become too large.

As it is shown in Fig. 11 and 12, the edges that could be extracted from the images are not identical due to higher detail in the scanned images, noise in the registered hyperspectral images, and changes introduced to the images during registration. However, the salient structures that are present in the images could all be extracted. In Fig. 11, the structures of the face are clearly visible and in Fig. 12, identical structures could be extracted in the border regions. Both figures show combinations of the edge images. Edges in the ground truth image are combined with edges from the RC or MI registered image respectively by overlaying the edge images and color encoding the edges. In the combined edge images, red edges represent edges from the reference image, blue edges represent edges from the registered image and green edges represent overlapping edges from the reference and registered images. It can be seen that many of the extracted salient structures directly overlap and may thus be considered as correctly registered. However, especially in the face images in Fig. 11, there are some regions where the images do not appear to be correctly registered. For example, the mouth and the structure on the right border seem to be incorrectly registered.

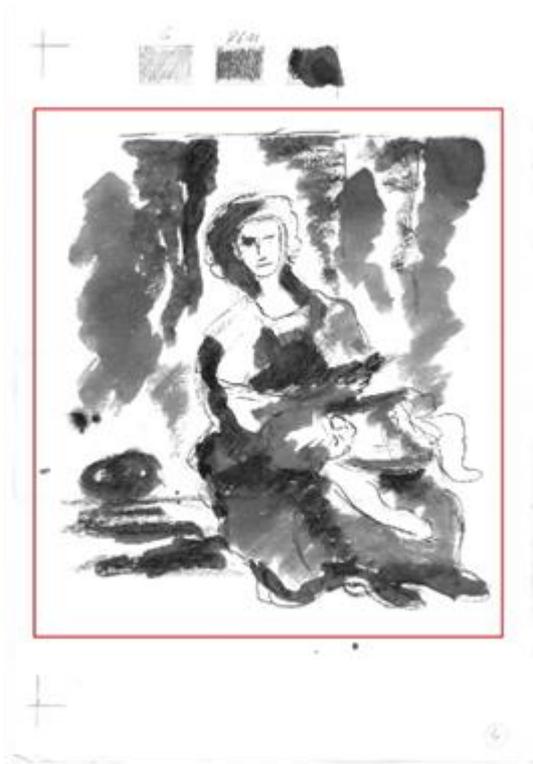
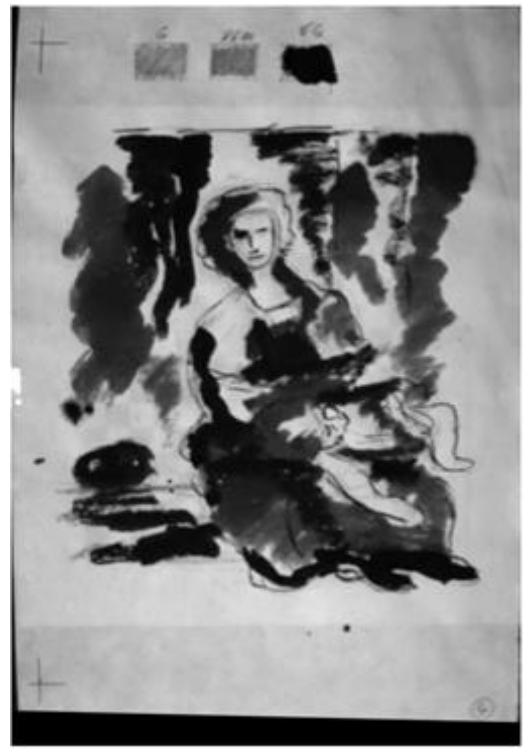

(a)                                                (b)

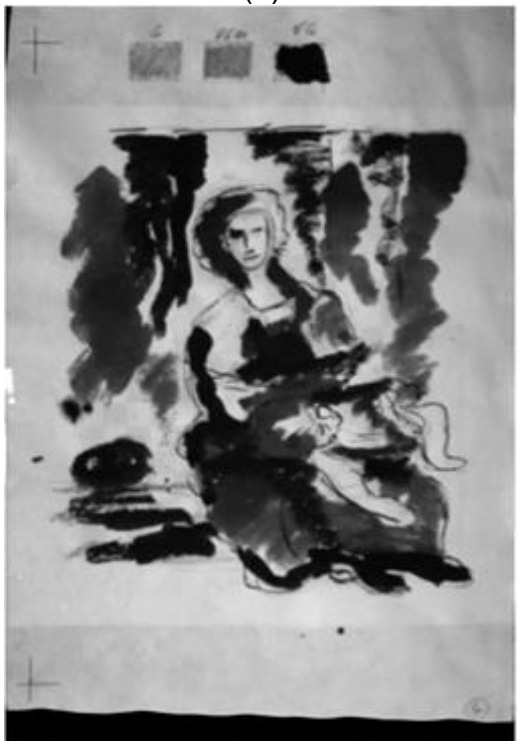
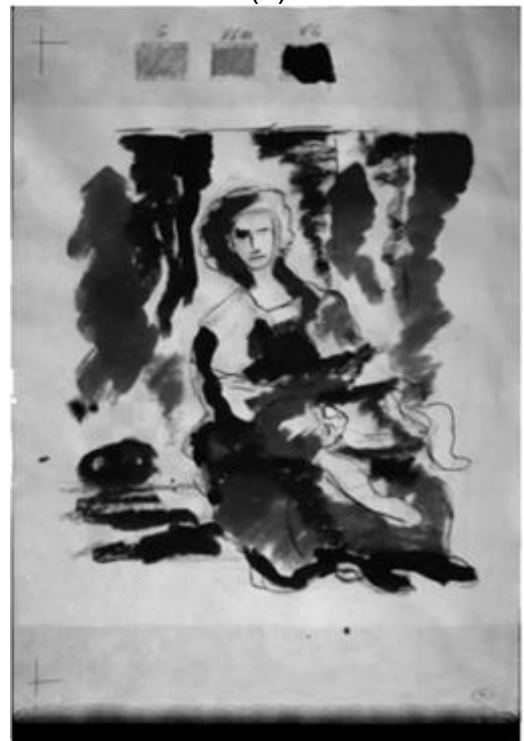

(c)                                                (d)

Figure 10: Input and output of the registration algorithm: (a) Reference image with region of interest; (b) Preregistered source image; (c) MI registration result; (d) RC registration result.

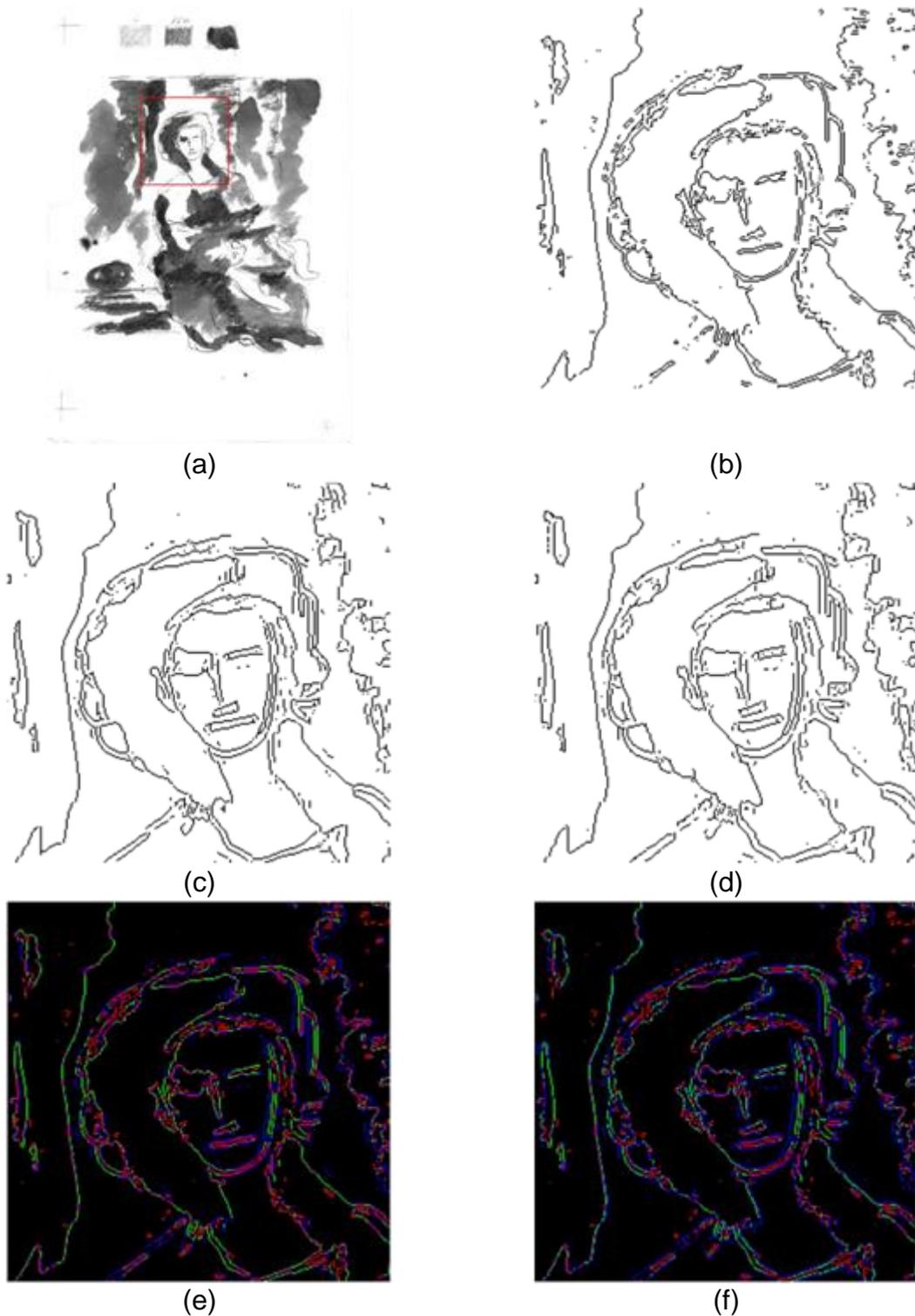

Figure 11: Edge images for the region with many details: (a) Reference image with extracted region; (b) edges in the reference image; (c) edges in the RC registered image; (d) edges in the MI registered image; (e) Combined edges from reference and RC registered image; (f) Combined edges from reference and MI registered image. In (e) and (f), red represents edges from the reference image, blue represents edges from the registered image and green represents overlapping edges from both images.

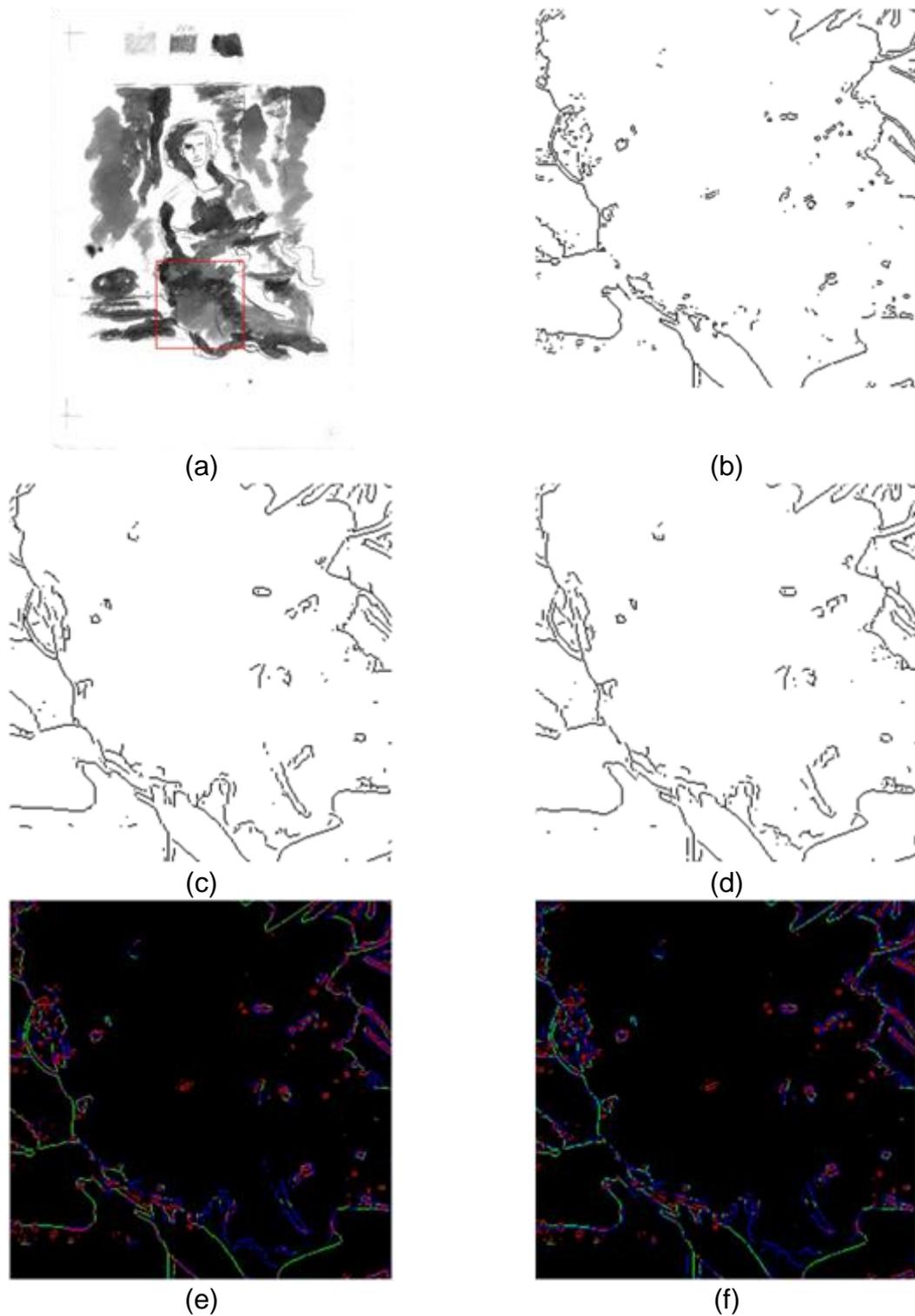

Figure 12: Edge images for the region with few details: (a) Reference image with extracted region; (b) edges in the reference image; (c) edges in the RC registered image; (d) edges in the MI registered image; (e) Combined edges from reference and RC registered image; (f) Combined edges from reference and MI registered image. In (e) and (f), red represents edges from the reference image, blue represents edges from the registered image and green represents overlapping edges from both images.

Results for the computation of the DICE similarity coefficient for the full area and both additional image regions can be seen in table 1. For the full area, our approach for the RC as well as the MI based registration shows a high improve in the regional overlap and both methods reach an equally high DSC score of 0.97. Both methods also provide good results in the high detailed image region and the low detailed image region. Here, the RC based methods shows a slight advantage in the high detailed region and the MI based method shows a slight advantage in the low detailed region.

| Image | no registration | RC registered | MI registered |
|---|---|---|---|
| Full area | 0.75 | 0.97 | 0.97 |
| High details | 0.64 | 0.95 | 0.94 |
| Low details | 0.41 | 0.91 | 0.92 |

Table 1: DICE similarity coefficient (DSC) for the different regions of interest (ROI) registration.

## 3 Discussion and Conclusion

In this work, we conducted a survey on different registration algorithms and investigated their suitability for hyperspectral historical image registration applications.

After the evaluation of different algorithms, we choose an intensity based registration algorithm with a curved transformation model. For the transformation model, we select cubic B-splines since they should be capable to cope with all non-rigid deformations in our hyperspectral images. For the similarity measure, we choose to evaluate a residual complexity based method against a mutual information based method. Both measures should be able to handle all difficulties, e.g., capture range, non-stationary and spatially varying intensity distortions or multi-modality that occur in our application. Further, we apply a pre-registration before the RC and MI based registration.

As we showed in section 2, the registration using RC as similarity measure as well as the registration using the MI as similarity measure perform equally well in the region of interest of our application. In addition, both algorithms perform well in the two selected regions where one shows many and the other few details. Here, the method using RC performs slightly better in the high detailed region and the MI performs slightly better in the low detailed region. One possibility for this behavior might be that the RC uses a frequency encoding of the images to measure their similarity and therefore benefits from a higher amount of details, i.e., frequencies, in the images. Therefore, it might be advantageous to apply a registration using the RC as similarity measures for images that show a high amount of details and the MI otherwise.

**References**

[Ade84]   Joan M, Adelson Edward H and Anderson, Charles H and Bergen, James R and Burt, Peter J and Ogden. "Pyramid methods in image processing." RCA engineer, 6th ser., 29 (1984): 33-41.

[Baj89]   Bajcsy, Ruzena and Kovacic, Stane. "Multiresolution elastic matching." Computer vision, graphics, and image processing, 1st ser., 46 (1989): 1-21.

[Bha01]   Bharatha, Aditya and Hirose, Masanori and Hata, Nobuhiko and Warfield, Simon K and Ferrant, Matthieu and Zou, Kelly H and Suarez-Santana, Eduardo and Ruiz-Alzola, Juan and D'amico, Anthony and Cormack, Robert A and others. "Evaluation of three-dimensional finite element-based deformable registration of pre-and intraoperative prostate imaging." Medical physics, 12th ser., 28 (2001): 2551-560.

[Chr01]   Christensen, Gary E and Johnson, Hans J. "Consistent image registration." IEEE transactions on medical imaging, 7th ser., 20 (2001): 568-82.

[Chr03]   Christensen, Gary E and Johnson, Hans J. "Invertibility and transitivity analysis for nonrigid image registration." Journal of electronic imaging, 1st ser., 12 (2003): 106-17.

[Chr06]   Christensen, Gary E and Geng, Xiujuan and Kuhl, Jon G and Bruss, Joel and Grabowski, Thomas J and Pirwani, Imran A and Vannier, Michael W and Allen, John S and Damasio, Hanna. "Introduction to the non-rigid image registration evaluation project (NIREP)." In International Workshop on Biomedical Image Registration, 128-35. Springer, 2006.

[Col95]   Collignon, Andre and Maes, Frederik and Delaere, Dominique and Vandermeulen, Dirk and Suetens, Paul and Marchal, Guy. "Automated multi-modality image registration based on information theory." In Information processing in medical imaging, 263-74. Vol. 3. Series 6. 1995.

[Cos14]   Cosentino, Antonino. "Identification of pigments by multispectral imaging; a flowchart method." Heritage Science, 1st ser., 2 (2014): 8.

[Cru14]   Crum, William R and Hartkens, Thomas and Hill, DLG. "Non-rigid image registration: theory and practice." The British Journal of Radiology, 2014.

[D'A03]   D'agostino, Emiliano and Maes, Frederik and Vandermeulen, Dirk and Suetens, Paul. "A viscous fluid model for multimodal non-rigid image registration using mutual information." Medical image analysis, 4th ser., 7 (2003): 565-75.

| | |
|---|---|
| [Dan11] | Danilchenko, Andrei and Fitzpatrick, J Michael. "General approach to first-order error prediction in rigid point registration." IEEE transactions on medical imaging, 3rd ser., 30 (2011): 679-93. |
| [Dav17] | Davari, A.; Häberle, A.; Christlein, V.; Maier, A.; Riess, C.: "Sketch Layer Separation in Multi-Spectral Historical Document Images." Pattern Recognition Lab, Friedrich-Alexander University, Tech. Rep. (2017). |
| [Dic45] | Dice, Lee R. "Measures of the amount of ecologic association between species." Ecology, 3rd ser., 26 (1945): 297-302. |
| [Fit98] | Fitzpatrick, J Michael and West, Jay B and Maurer, Calvin R. "Predicting error in rigid-body point-based registration." IEEE transactions on medical imaging, 5$^{th}$ ser., 17 (1998): 694-702. |
| [Fit00] | Fitzpatrick, J Michael and Sonka, Milan. Medical Image Processing and Analysis. SPIE Press, 2000. |
| [Gol65] | Goldstein, Herbert and Poole, Charles and Safko, John. Classical mechanics. AAPT, 2002. |
| [Haj95] | Hajnal, Joseph V and Saeed, Nadeem and Soar, Elaine J and Oatridge, Angela and Young, Ian R and Bydder, Graeme M. "A registration and interpolation procedure for subvoxel matching of serially acquired MR images." Journal of computer assisted tomography, 2nd ser., 19 (1995): 289-96. |
| [Har03] | Hartley, Richard and Zisserman, Andrew. Multiple view geometry in computer vision. Cambridge university press, 2003. |
| [Kle08] | Klein, Stefan and van der Heide, Uulke A and Lips, Irene M and van Vulpen, Marco and Staring, Marius and Pluim, Josien PW. "Automatic segmentation of the prostate in 3D MR images by atlas matching using localized mutual information." Medical physics, 4th ser., 35 (2008): 1407-417. |
| [Kul51] | Kullback, Solomon and Leibler, Richard A. "On information and sufficiency." The annals of mathematical statistics, 1st ser., 22 (1951): 79-86. |
| [Loe10] | Loeckx, Dirk and Slagmolen, Pieter and Maes, Frederik and Vandermeulen, Dirk and Suetens, Paul. "Nonrigid image registration using conditional mutual information." IEEE transactions on medical imaging, 1st ser., 29 (2010): 19-29. |
| [Mai98] | Maintz, JB Antoine and Viergever, Max A. "A survey of medical image registration." Medical image analysis, 1st ser., 2 (1998): 1-36. |
| [Mar12] | Markelj, Primoz and Tomazevic, Dejan and Likar, Bostjan and Pernus, Franjo. "A review of 3D/2D registration methods for image-guided interventions." Medical image analysis, 3rd ser., 16 (2012): 642-61. |
| [Mid16a] | Middleton Spectral Vision. VNIR Cameras. Accessed July 2016. http://www.middletonspectral.com/products/hyperspectral-components-systems/hyperspectral-cameras/vnir-cameras/. |
| [Mid16b] | Middleton Spectral Vision. What are the key components of a hyperspectral system? July 2016. http://www.middletonspectral.com/what-are-the-key-components-of-a-hyperspectral-system/. |
| [Mid16c] | Middleton Spectral Vision. What is hyperspectral imaging? Accessed July 2016. http:// www.middletonspectral.com/what-is-hyperspectral-imaging/. |
| [Myr10] | Myronenko, A.; Song, X.: Intensity-Based Image Registration by Minimizing Residual Complexity, IEEE Transactions on Medical Imaging, Bd. 29, Nr. 11, Nov 2010, S. 1882–1891. |
| [Myr16] | Myronenko, Andriy and Song, Xubo. "Intensity-based image registration by minimizing residual complexity." IEEE Transactions on Medical Imaging, 11th ser., 29 (2010): 1882-891. |
| [Oli14] | Oliveira, Francisco PM and Tavares, Joao Manuel RS. "Medical image registration: a review." Computer methods in biomechanics and biomedical engineering, 2nd ser., 17 (2014): 73-93. |
| [Pre07] | Press, W. H.: Numerical recipes 3rd edition: The art of scientific computing, Cambridge university press, 2007. |
| [Roc98] | Roche, Alexis and Malandain, Gregoire and Pennec, Xavier and Ayache, Nicholas. "The correlation ratio as a new similarity measure for multimodal image registration." International Conference on Medical Image Computing and Computer-Assisted Intervention, 1998, 1115-124. |
| [Roc99] | Roche Alexis and Malandain, Gregoire and Ayache, Nicholas,"Unifying maximum likelihood approaches in medical image registration " PhD diss., Inria, 1999 |
| [Rue99] | Rueckert, Daniel and Sonoda, Luke I and Hayes, Carmel and Hill, Derek LG and Leach, Martin O and Hawkes, David J. "Nonrigid registration using free-form deformations: application to breast MR images." IEEE transactions on medical imaging, 8th ser., 18 (1999): 712-21. |
| [Sch01] | Schnabel, Julia A. 'A generic framework for non-rigid registration based on non-uniform multi-level free-form deformations'. In International Conference on Medical Image Computing and Computer-Assisted Intervention. |
| [Sha01] | Shannon, Claude Elwood. "A mathematical theory of communication." ACM SIGMOBILE Mobile Computing and Communications Review, 1st ser., 5 (2001): 3-55. |
| [Sob68] | Sobel, Irwin and Feldman, Gary. "A 3x3 isotropic gradient operator for image processing." A talk at the Stanford Artificial Project in, 1968, 271-72. |
| [Stu99] | Studholme, Colin and Hill, Derek LG and Hawkes, David J. "An overlap invariant entropy measure of 3D medical image alignment." Pattern recognition, 1st ser., 32 (1999): 71-86. |
| [Stu06] | Studholme, Colin and Drapaca, Corina and Iordanova, Bistra and Cardenas, Valerie. "Deformation-based mapping of volume change from serial brain MRI in the presence of local tissue contrast change." IEEE transactions on Medical Imaging, 5th ser., 25 (2006): 626-39. |
| [Urs07] | Urschler, Martin. 'A framework for comparison and evaluation of nonlinear intra-subject image registration algorithms'. In IJ-2007 MICCAI Open Science Workshop, 2007 |
| [Ver07] | Vercauteren, Tom. 'Non-parametric diffeomorphic image registration with the demons algorithm'. In International Conference on Medical Image Computing and Computer-Assisted Intervention, 319–326. Springer, 2007. |
| [Vio97] | Viola, Paul and Wells III, William M. "Alignment by maximization of mutual information." International journal of computer vision, 2nd ser., 24 (1997): 137-54. |
| [Wan05] | Wang, He and Dong, Lei and O'Daniel, Jennifer and Mohan, Radhe and Garden, Adam S and Ang, K Kian and Kuban, Deborah A and Bonnen, Mark and Chang, Joe Y and Cheung, Rex. "Validation of an accelerated 'demons' algorithm for deformable image registration in radiation therapy." Physics in medicine and biology, 12th ser., 50 (2005): 2887-905. |
| [Zit03] | Zitova, Barbara and Flusser, Jan. "Image registration methods: a survey." Image and vision computing, 11th ser., 21 (2003): 977-1000. |